\documentclass[]{article} 
\usepackage{proceed2e}

\usepackage{graphicx}
\usepackage{times}
\usepackage{amsmath}
\usepackage{amssymb}
\usepackage{paralist}
\usepackage{url}
\usepackage[round,sort&compress]{natbib}
\usepackage[all]{xy}

\newcommand{\defoccur}[1]{\textsl{#1}}

\usepackage{xspace}
\def\onedot{\ifx\@let@token.\else.\null\fi\xspace}
\def\eg{{e.g}\onedot} 
\def\ie{{i.e}\onedot} 
 
\def\etc{{etc}\onedot}
\def\vs{{vs}\onedot}

\def\etal{{et al}\onedot}

\begin{document}

\title{Video In Sentences Out}

\author{Andrei Barbu,$^{a}$\thanks{%
    \hspace{1ex}Corresponding author. Email: {\small\texttt{andrei@0xab.com}}.}$\;$
Alexander Bridge,$^{a}$
Zachary Burchill,$^{a}$
Dan Coroian,$^{a}$
Sven Dickinson,$^{b}$\\
Sanja Fidler,$^{b}$
Aaron Michaux,$^{a}$
Sam Mussman,$^{a}$
Siddharth Narayanaswamy,$^{a}$
Dhaval Salvi,$^{c}$\\
Lara Schmidt,$^{a}$
Jiangnan Shangguan,$^{a}$
Jeffrey Mark Siskind,$^{a}$
Jarrell Waggoner,$^{c}$
Song Wang,$^{c}$\\
Jinlian Wei,$^{a}$
Yifan Yin,$^{a}$ and
Zhiqi Zhang$^{c}$}

\date{\begin{tabular}[t]{l}
$^{a}$School of Electrical \& Computer Engineering, Purdue University,
West Lafayette, IN, USA\\
$^{b}$Department of Computer Science, University of Toronto, Toronto,
CA\\
$^{c}$Department of Computer Science \& Engineering, University of
South Carolina, Columbia, SC, USA
\end{tabular}}

\maketitle

\let\thefootnote\relax\footnotetext{Additional images and videos as well as all
  code and datasets are available at
  \url{http://engineering.purdue.edu/~qobi/arxiv2012b}.}

\begin{abstract}
  We present a system that produces sentential descriptions of video: who did
  what to whom, and where and how they did it.
  Action class is rendered as a verb, participant objects as noun phrases,
  properties of those objects as adjectival modifiers in those noun phrases,
  spatial relations between those participants as prepositional phrases, and
  characteristics of the event as prepositional-phrase adjuncts and adverbial
  modifiers.
  Extracting the information needed to render these linguistic entities requires
  an approach to event recognition that recovers object tracks, the
  track-to-role assignments, and changing body posture.
\end{abstract}

\section{Introduction}

We present a system that produces sentential descriptions of short video clips.
These sentences describe \emph{who} did \emph{what} to \emph{whom}, and
\emph{where} and \emph{how} they did it.
This system not only describes the observed action as a verb, it also describes
the participant objects as noun phrases, properties of those objects as
adjectival modifiers in those noun phrases, the spatial relations between those
participants as prepositional phrases, and characteristics of the event as
prepositional-phrase adjuncts and adverbial modifiers.
It incorporates a vocabulary of 118~words: 1~coordination, 48~verbs, 24~nouns,
20~adjectives, 8~prepositions, 4~lexical prepositional phrases, 4~determiners,
3~particles, 3~pronouns, 2~adverbs, and 1~auxiliary, as illustrated in
Table~\ref{tab:vocabulary}.

\begin{table}
  \begin{scriptsize}
    \begin{tabbing}
      \textbf{coordination}: \=
      \emph{and}\\
      \textbf{verbs}:\>
      \emph{approached},
      \emph{arrived},
      \emph{attached},
      \emph{bounced},
      \emph{buried},
      \emph{carried},
      \emph{caught},\\
      \>\emph{chased},
      \emph{closed},
      \emph{collided},
      \emph{digging},
      \emph{dropped},
      \emph{entered},
      \emph{exchanged},\\
      \>\emph{exited},
      \emph{fell},
      \emph{fled},
      \emph{flew},
      \emph{followed},
      \emph{gave},
      \emph{got},
      \emph{had},
      \emph{handed},
      \emph{hauled},
      \emph{held},\\
      \>\emph{hit},
      \emph{jumped},
      \emph{kicked},
      \emph{left},
      \emph{lifted},
      \emph{moved},
      \emph{opened},
      \emph{passed},
      \emph{picked},\\
      \>\emph{pushed},
      \emph{put},
      \emph{raised},
      \emph{ran},
      \emph{received},
      \emph{replaced},
      \emph{snatched},
      \emph{stopped},\\
      \>\emph{threw},
      \emph{took},
      \emph{touched},
      \emph{turned},
      \emph{walked},
      \emph{went}\\
      \textbf{nouns}:\>
      \emph{bag},
      \emph{ball},
      \emph{bench},
      \emph{bicycle},
      \emph{box},
      \emph{cage},
      \emph{car},
      \emph{cart},
      \emph{chair},
      \emph{dog},
      \emph{door},\\
      \>\emph{ladder},
      \emph{left},
      \emph{mailbox},
      \emph{microwave},
      \emph{motorcycle},
      \emph{object},
      \emph{person},
      \emph{right},\\
      \>\emph{skateboard},
      \emph{SUV},
      \emph{table},
      \emph{tripod},
      \emph{truck}\\
      \textbf{adjectives}:\>
      \emph{big},
      \emph{black},
      \emph{blue},
      \emph{cardboard},
      \emph{crouched},
      \emph{green},
      \emph{narrow},
      \emph{other},
      \emph{pink},\\
      \>\emph{prone},
      \emph{red},
      \emph{short},
      \emph{small},
      \emph{tall},
      \emph{teal},
      \emph{toy},
      \emph{upright},
      \emph{white},
      \emph{wide},
      \emph{yellow}\\
      \textbf{prepositions}:\>
      \emph{above},
      \emph{because},
      \emph{below},
      \emph{from},
      \emph{of},
      \emph{over},
      \emph{to},
      \emph{with}\\
      \textbf{lexical PPs}:\>
      \emph{downward},
      \emph{leftward},
      \emph{rightward},
      \emph{upward}\\
      \textbf{determiners}:\>
      \emph{an},
      \emph{some},
      \emph{that},
      \emph{the}\\
      \textbf{particles}:\>
      \emph{away},
      \emph{down},
      \emph{up}\\
      \textbf{pronouns}:\>
      \emph{itself},
      \emph{something},
      \emph{themselves}\\
      \textbf{adverbs}:\>
      \emph{quickly},
      \emph{slowly}\\
      \textbf{auxiliary}:\>
      \emph{was}\\
    \end{tabbing}
  \end{scriptsize}
  \vspace{-3ex}
  \caption{The vocabulary used to generate sentential descriptions of video.}
  \label{tab:vocabulary}
  \vspace{-2ex}
\end{table}

Production of sentential descriptions requires recognizing the primary action
being performed, because such actions are rendered as verbs and verbs serve as
the central scaffolding for sentences.
However, event recognition alone is insufficient to generate the remaining
sentential components.
One must recognize object classes in order to render nouns.
But even object recognition alone is insufficient to generate meaningful
sentences.
One must determine the \defoccur{roles} that such objects play in the event.
The \defoccur{agent}, \ie\ the doer of the action, is typically rendered as the
sentential subject while the \defoccur{patient}, \ie\ the affected object, is
typically rendered as the direct object.
Detected objects that do not play a role in the observed event, no matter how
prominent, should not be incorporated into the description.
This means that one cannot use common approaches to event recognition, such as
spatiotemporal bags of words \citep{Laptev2007, Niebles2008, Scovanner2007},
spatiotemporal volumes \citep{Blank2005, Laptev2008, Rodriguez2008}, and tracked
feature points \citep{Liu2009, Schuldt2004, Wang2009} that do not determine the
class of participant objects and the roles that they play.
Even combining such approaches with an object detector would likely detect
objects that don't participate in the event and wouldn't be able to determine
the roles that any detected objects play.

Producing elaborate sentential descriptions requires more than just event
recognition and object detection.
Generating a noun phrase with an embedded prepositional phrase, such as
\emph{the person to the left of the bicycle}, requires determining spatial
relations between detected objects, as well as knowing which of the two
detected objects plays a role in the overall event and which serves just to
aid generation of a referring expression to help identify the event
participant.
Generating a noun phrase with adjectival modifiers, such as \emph{the red ball},
not only requires determining the properties, such as color, shape, and size,
of the observed objects, but also requires determining whether such
descriptions are necessary to help disambiguate the referent of a noun phrase.
It would be awkward to generate a noun phrase such as \emph{the big tall
  wide red toy cardboard trash can} when \emph{the trash can} would suffice.
Moreover, one must track the participants to determine the speed and direction
of their motion to generate adverbs such as \emph{slowly} and prepositional
phrases such as \emph{leftward}.
Further, one must track the identity of multiple instances of the same object
class to appropriately generate the distinction between \emph{Some person hit
  some other person} and \emph{The person hit themselves}.

A common assumption in Linguistics \citep{Jackendoff83, Pinker89} is that verbs
typically characterize the interaction between event participants in terms of
the gross changing motion of these participants.
Object class and image characteristics of the participants are believed to be
largely irrelevant to determining the appropriate verb label for an action
class.
Participants simply fill roles in the spatiotemporal structure of the action
class described by a verb.
For example, an event where one participant (the agent) \emph{picks up}
another participant (the patient) consists of a sequence of two sub-events,
where during the first sub-event the agent moves towards the patient while the
patient is at rest and during the second sub-event the agent moves together
with the patient away from the original location of the patient.
While determining whether the agent is a \emph{person} or a \emph{cat}, and
whether the patient is a \emph{ball} or a \emph{cup}, is necessary to generate
the noun phrases incorporated into the sentential description, such information
is largely irrelevant to determining the verb describing the action.
Similarly, while determining the shapes, sizes, colors, textures, \etc{} of the
participants is necessary to generate adjectival modifiers, such information is
also largely irrelevant to determining the verb.
Common approaches to event recognition, such as spatiotemporal bags of words,
spatiotemporal volumes, and tracked feature points, often achieve high accuracy
because of correlation with image or video properties exhibited by a particular
corpus.
These are often artefactual, not defining properties of the verb meaning
(\eg\ recognizing \emph{diving} by correlation with \emph{blue} since it
`happens in a pool' \cite[p.\ 2002]{Liu2009} or confusing \emph{basketball} and
\emph{volleyball} `because most of the time the $[\ldots]$ sports use very
similar courts' \cite[p.\ 506]{Ikizler2010}).

\section{The mind's eye corpus}

Many existing video corpora used to evaluate event recognition are ill-suited
for evaluating sentential descriptions.
For example, the \textsc{Weizmann} dataset \citep{Blank2005} and the \textsc{kth}
dataset \citep{Schuldt2004} depict events with a single human participant, not
ones where people interact with other people or objects.
For these datasets, the sentential descriptions would contain no information
other than the verb, \eg\ \emph{The person jumped}.
Moreover, such datasets, as well as the \textsc{Sports Actions} dataset
\citep{Rodriguez2008} and the \textsc{Youtube} dataset \citep{Liu2009}, often make
action-class distinctions that are irrelevant to the choice of verb,
\eg\ \texttt{wave1} \vs\ \texttt{wave2}, \texttt{jump} \vs\ \texttt{pjump},
\texttt{Golf-Swing-Back} \vs\ \texttt{Golf-Swing-Front}
\vs\ \texttt{Golf-Swing-Side}, \texttt{Kicking-Front}
\vs\ \texttt{Kicking-Side}, \texttt{Swing-Bench}
\vs\ \texttt{Swing-SideAngle}, and \texttt{golf\_swing} \vs\
\texttt{tennis\_swing} \vs\ \texttt{swing}
Other datasets, such as the \textsc{Ballet} dataset \citep{Wang2009} and
the \textsc{ucf50} dataset \citep{Liu2009}, depict larger-scale activities that
bear activity-class names that are not well suited to sentential
description, \eg\ \texttt{Basketball}, \texttt{Billiards},
\texttt{BreastStroke}, \texttt{CleanAndJerk}, \texttt{HorseRace},
\texttt{HulaHoop}, \texttt{MilitaryParade}, \texttt{TaiChi}, and \texttt{YoYo}.

The year-one (Y1) corpus produced by DARPA for the Mind's Eye program, however,
was specifically designed to evaluate sentential description.
This corpus contains two parts: the development corpus, C-D1, which we use
solely for training, and the evaluation corpus, C-E1, which we use solely for
testing.
Each of the above is further divided into four sections to support the four
task goals of the Mind's Eye program, namely recognition, description, gap
filling, and anomaly detection.
In this paper, we use only the recognition and description portions and apply
our entire sentential-description pipeline to the combination of these
portions.
While portions of C-E1 overlap with C-D1, \textbf{\emph{in this paper we
    train our methods solely on C-D1 and test our methods solely on the portion
    of C-E1 that does not overlap with C-D1}}.

Moreover, a portion of the corpus was synthetically generated by a variety of
means: computer graphics driven by motion capture, pasting foregrounds
extracted from green screening onto different backgrounds, and intensity
variation introduced by postprocessing.
\textbf{\emph{In this paper, we exclude all such synthetic video from our test
    corpus.}}
Our training set contains 3480 videos and our test set 749 videos.
These videos are provided at 720p@30fps and range from 42 to 1727 frames in
length, with an average of 435 frames.

The videos nominally depict 48 distinct verbs as listed in
Table~\ref{tab:vocabulary}.
However, the mapping from videos to verbs is not one-to-one.
Due to polysemy, a verb may describe more than one action class,
\eg\ \emph{leaving an object on the table} \vs\ \emph{leaving the scene}.
Due to synonymy, an action class may be described by more than one verb,
\eg\ \emph{lift} \vs\ \emph{raise}.
An event described by one verb may contain a component action described by a
different verb, \eg\ \emph{picking up an object} \vs\ \emph{touching an object}.
Many of the events are described by the combination of a verb with other
constituents, \eg\ \emph{have a conversation} \vs\ \emph{have a heart attack}.
And many of the videos depict metaphoric extensions of verbs, \eg\ \emph{take
  a puff on a cigarette}.
Because the mapping from videos to verbs is subjective, the corpus comes
labeled with DARPA-collected human judgments in the form of a single
present/absent label associated with each video paired with each of the 48
verbs, gathered using Amazon Mechanical Turk.
We use these labels for both training and testing as described later.

\section{Overall system architecture}

\begin{figure}
\begin{center}
\includegraphics[width=0.99\columnwidth]{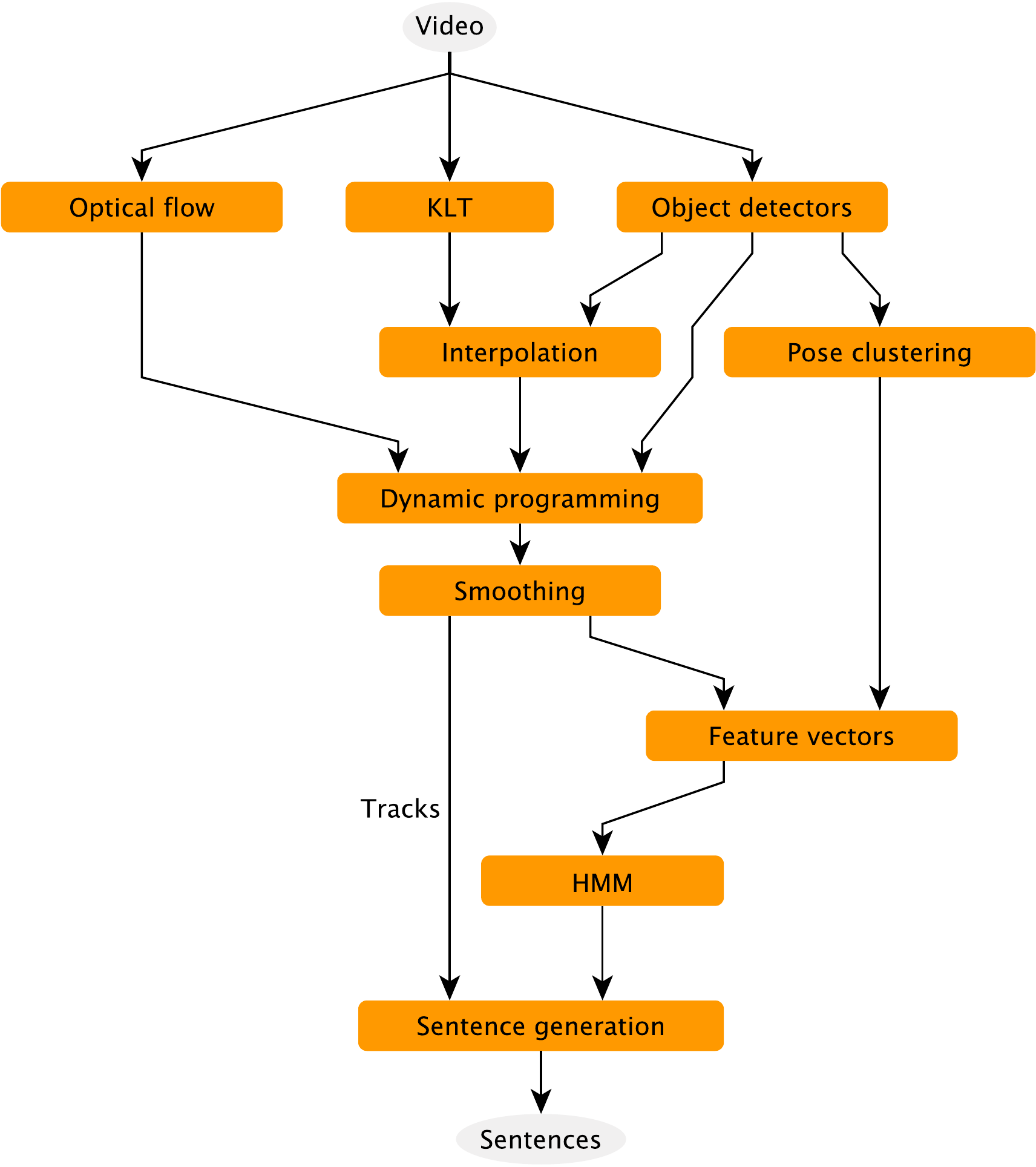}
\par\vspace*{2ex}
\end{center}
\caption{The overall architecture of our system for producing sentential
  descriptions of video.}
\label{fig:architecture}
\par\vspace*{2ex}
\end{figure}

The overall architecture of our system is depicted in
Fig.~\ref{fig:architecture}.
We first apply detectors \citep{Felzenszwalb2010b, Felzenszwalb2010a} for each
object class on each frame of each video.
These detectors are biased to yield many false positives but few false
negatives.
The Kanade-Lucas-Tomasi (KLT) \citep{shi1994, tomasi1991} feature tracker is then
used to project each detection five frames forward to augment the set of
detections and further compensate for false negatives in the raw detector
output.
A dynamic-programming algorithm \citep{Viterbi1971} is then used to select an
optimal set of detections that is temporally coherent with optical flow,
yielding a set of object tracks for each video.
These tracks are then smoothed and used to compute a time-series of feature
vectors for each video to describe the relative and absolute motion of event
participants.
The person detections are then clustered based on part displacements to derive a
coarse measure of human body posture in the form of a body-posture codebook.
The codebook indices of person detections are then added to the feature vector.
Hidden Markov Models (HMMs) are then employed as time-series classifiers to
yield verb labels for each video \citep{Siskind1996, Starner98, Wang2009,
  Xu2002, Xu2005}, together with the object tracks of the participants in the
action described by that verb along with the roles they play.
These tracks are then processed to produce nouns from object classes,
adjectives from object properties, prepositional phrases from spatial
relations, and adverbs and prepositional-phrase adjuncts from track properties.
Together with the verbs, these are then woven into grammatical sentences.
We describe each of the components of this system in detail below:
the object detector and tracker in Section~\ref{sec:tracking}, the
body-posture clustering and codebook in Section~\ref{sec:posture}, the event
classifier in Section~\ref{sec:classifier}, and the sentential-description
component in Section~\ref{sec:description}.

\subsection{Object detection and tracking}
\label{sec:tracking}

We employ detection-based tracking as described in Section~2 of a parallel
submission (id: 568)
In detection-based tracking an object detector is applied to each frame of a
video to yield a set of candidate detections which are composed into tracks by
selecting a single candidate detection from each frame that maximizes temporal
coherency of the track.
Felzenszwalb \etal\ detectors are used for this purpose.
Detection-based tracking requires biasing the detector to have high recall at
the expense of low precision to allow the tracker to select boxes to yield a
temporally coherent track.
This is done by depressing the acceptance thresholds.
To prevent massive over-generation of false positives, which would severely
impact run time, we limit the number of detections produced per-frame to~12.

Two practical issues arise when depressing acceptance thresholds.
First, it is necessary to reduce the degree of non-maximal suppression
incorporated in the Felzenszwalb \etal\ detectors.
Second, with the star detector \citep{Felzenszwalb2010a}, one can simply
decrease the single trained acceptance threshold to yield more detections with
no increase in computational complexity.
However, we prefer to use the star cascade detector \citep{Felzenszwalb2010b} as
it is far faster.
With the star cascade detector, though, one must also decrease the trained
root- and part-filter thresholds to get more detections.
Doing so, however, defeats the computational advantage of the cascade and
significantly increases detection time.
We thus train a model for the star detector using the standard procedure on
human-annotated training data, sample the top detections produced by this model
with a decreased acceptance threshold, and train a model for the star cascade
detector on these samples.
This yields a model that is almost as fast as one trained by the star cascade
detector on the original training samples but with the desired bias in
acceptance threshold.

The Y1 corpus contains approximately 70 different object classes that play a
role in the depicted events.
Many of these, however, cannot be reliably detected with the Felzenszwalb
 \etal\ detectors that we use.
We trained models for 25 object classes that can be reliably detected, as
listed in Table~\ref{tab:objects}.
These object classes account for over 90\% of the event participants.
Person models were trained with approximately 2000 human-annotated positive
samples from C-D1 while nonperson models were trained with approximately 1000
such samples.
For each positive training sample, two negative training samples were randomly
generated from the same frame constrained to not overlap substantially with the
positive samples.
We trained three distinct person models to account for body-posture variation
and pool these when constructing person tracks.
The detection scores were normalized for such pooled detections by a per-model
offset computed as follows:
A (50 bin) histogram was computed of the scores of the top detection in each
frame of a video.
The offset is then taken to be the minimum of the value that maximizes the
between-class variance \citep{Otsu1979} when bipartitioning this histogram and
the trained acceptance threshold offset by a fixed, but small, amount (0.4).

\begin{table*}
  \begin{center}
    \begin{tabular}{ll}
      (a)&
      \scalebox{0.8}
	       {\begin{tabular}{lllll}
		\texttt{bag}$\mapsto$\emph{bag}&
		\texttt{car}$\mapsto$\emph{car}&
		\texttt{door}$\mapsto$\emph{door}&
		\texttt{person}$\mapsto$\emph{person}&
		\texttt{suv}$\mapsto$\emph{SUV}\\
		\texttt{bench}$\mapsto$\emph{bench}&
		\texttt{cardboard-box}$\mapsto$\emph{box}&
		\texttt{ladder}$\mapsto$\emph{ladder}&
		\texttt{person-crouch}$\mapsto$\emph{person}&
		\texttt{table}$\mapsto$\emph{table}\\
		\texttt{bicycle}$\mapsto$\emph{bicycle}&
		\texttt{cart}$\mapsto$\emph{cart}&
		\texttt{mailbox}$\mapsto$\emph{mailbox}&
		\texttt{person-down}$\mapsto$\emph{person}&
		\texttt{toy-truck}$\mapsto$\emph{truck}\\
		\texttt{big-ball}$\mapsto$\emph{ball}&
		\texttt{chair}$\mapsto$\emph{chair}&
		\texttt{microwave}$\mapsto$\emph{microwave}&
		\texttt{skateboard}$\mapsto$\emph{skateboard}&
		\texttt{tripod}$\mapsto$\emph{tripod}\\
		\texttt{cage}$\mapsto$\emph{cage}&
		\texttt{dog}$\mapsto$\emph{dog}&
		\texttt{motorcycle}$\mapsto$\emph{motorcycle}&
		\texttt{small-ball}$\mapsto$\emph{ball}&
		\texttt{truck}$\mapsto$\emph{truck}
		\end{tabular}}\\\hline
	       (b)&
	       \scalebox{0.8}
			{\begin{tabular}{lllll}
			 \texttt{cardboard-box}$\mapsto$\emph{cardboard}&
                         \texttt{person}$\mapsto$\emph{upright}&
                         \texttt{person-crouch}$\mapsto$\emph{crouched}&
                         \texttt{person-down}$\mapsto$\emph{prone}&
			 \texttt{toy-truck}$\mapsto$\emph{toy}
			 \end{tabular}}\\\hline
			(c)&
			\scalebox{0.8}
				 {\begin{tabular}{ll}
				  \texttt{big-ball}$\mapsto$\emph{big}&
				  \texttt{small-ball}$\mapsto$\emph{small}
				  \end{tabular}}
    \end{tabular}
  \end{center}
  \caption{Trained models for object classes and their mappings to
    (a)~nouns, (b)~restrictive adjectives, and (c)~size adjectives.}
  \label{tab:objects}
\end{table*}

We employed detection-based tracking for all 25 object models on all 749 videos
in our test set.
To prune the large number of tracks thus produced, we discard all tracks
corresponding to certain object models on a per-video basis: those that
exhibit high detection-score variance over the frames in that video as well as
those whose detection-score distributions are neither unimodal nor bimodal.
The parameters governing such pruning were determined solely on the training
set.
The tracks that remain after this pruning still account for over 90\% of the
event participants.

\subsection{Body-posture codebook}
\label{sec:posture}

We recognize events using a combination of the motion of the event
participants and the changing body posture of the human participants.
Body-posture information is derived using the part structure produced as a
by-product of the Felzenszwalb \etal\ detectors.
While such information is far noisier and less accurate than fitting precise
articulated models \citep{Andriluka2008, Bregler1997, GavrilaD95, Sigal2010,
  Yang2011} and appears unintelligible to the human eye, as shown in
Section~\ref{sec:classifier}, it suffices to improve event-recognition
accuracy.
Such information can be extracted from a large unannotated corpus far
more robustly than possible with precise articulated models.

\begin{figure*}
  \begin{center}
    \begin{tabular}{@{}c@{\hspace*{6pt}}c@{}}
      \begin{tabular}{@{}c@{\hspace*{2pt}}c@{\hspace*{2pt}}c@{\hspace*{2pt}}c@{\hspace*{2pt}}c@{}}
          \includegraphics[width=0.095\textwidth]{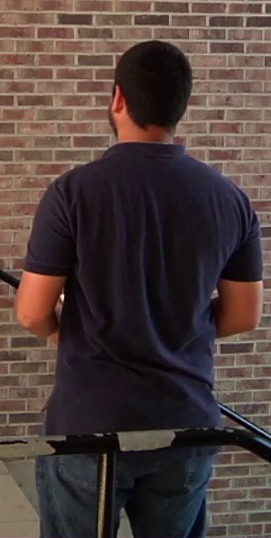}&
          \includegraphics[width=0.095\textwidth]{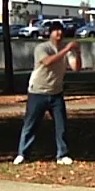}&
          \includegraphics[width=0.095\textwidth]{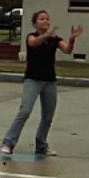}&
          \includegraphics[width=0.095\textwidth]{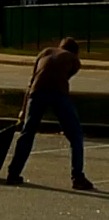}&
          \includegraphics[width=0.095\textwidth]{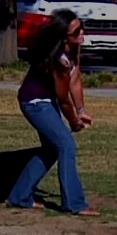}
      \end{tabular}
      &
      \begin{tabular}{@{}c@{\hspace*{2pt}}c@{\hspace*{2pt}}c@{\hspace*{2pt}}c@{\hspace*{2pt}}c@{}}
        \includegraphics[width=0.095\textwidth]{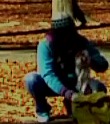}&
        \includegraphics[width=0.095\textwidth]{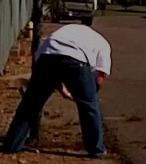}&
        \includegraphics[width=0.095\textwidth]{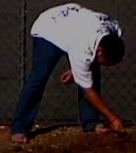}&
        \includegraphics[width=0.095\textwidth]{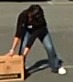}&
        \includegraphics[width=0.095\textwidth]{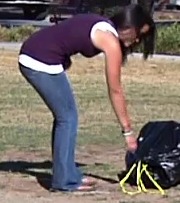}\\*[7pt]
        \includegraphics[width=0.095\textwidth]{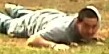}&
        \includegraphics[width=0.095\textwidth]{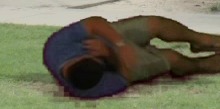}&
        \includegraphics[width=0.095\textwidth]{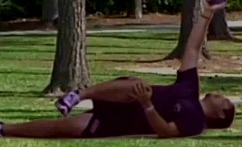}&
        \includegraphics[width=0.095\textwidth]{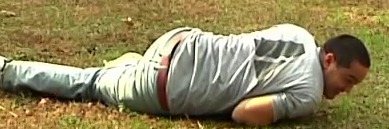}&
        \includegraphics[width=0.095\textwidth]{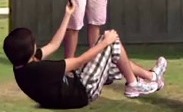}
      \end{tabular}
    \end{tabular}
  \end{center}
  \caption{Sample clusters from our body-posture codebook.}
  \label{fig:codebook}
  \vspace{-1ex}
\end{figure*}

Body-posture information is derived from part structure in two ways.
First, we compute a vector of part displacements, each displacement as a vector
from the detection center to the part center, normalizing these vectors to unit
detection-box area.
The time-series of feature vectors is augmented to includes these part
displacements and a finite-difference approximation of their temporal
derivatives as continuous features for person detections.
Second, we vector-quantize the part-displacement vector and include the
codebook index as a discrete feature for person detections.
Such pose features are included in the time-series on a per-frame basis.
The codebook is trained by running each pose-specific person detector on the
positive human-annotated samples used to train that detector and extract the
resulting part-displacement vectors.
We then pool the part-displacement vectors from the three pose-specific person
models and employ hierarchical $k$-means clustering using Euclidean distance
to derive a codebook of 49 clusters.
Fig.~\ref{fig:codebook} shows sample clusters from our codebook.
Codebook indices are derived using Euclidean distance from the means of these
clusters.

\subsection{Event classification}
\label{sec:classifier}

Our tracker produces one or more tracks per object class for each video.
We convert such tracks into a time-series of feature vectors.
For each video, one track is taken to designate the agent and another track (if
present) is taken to designate the patient.
During training, we manually specify the track-to-role mapping.
During testing, we automatically determine the track-to-role mapping by
examining all possible such mappings and selecting the one with the highest
likelihood \citep{Siskind1996}.

The feature vector encodes both the motion of the event participants and the
changing body posture of the human participants.
For each event participant in isolation we incorporate the following
single-track features:
\begin{compactenum}
\item $x$ and $y$ coordinates of the detection-box center
\item detection-box aspect ratio and its temporal derivative
\item magnitude and direction of the velocity of the detection-box center
\item magnitude and direction of the acceleration of the detection-box center
\item normalized part displacements and their temporal derivatives
\item object class (the object detector yielding the detection)
\item root-filter index
\item body-posture codebook index
\end{compactenum}
The last three features are discrete; the remainder are continuous.
For each pair of event participants we incorporate the following track-pair
features:
\begin{compactenum}
\item distance between the agent and patient detection-box centers and its
  temporal derivative
\item orientation of the vector from agent detection-box center to patient
  detection-box center
\end{compactenum}

Our HMMs assume independent output distributions for each feature.
Discrete features are modeled with discrete output distributions.
Continuous features denoting linear quantities are modeled with univariate
Gaussian output distributions, while those denoting angular quantities are
modeled with von~Mises output distributions.

For each of the 48 action classes, we train two HMMs on two different sets of
time-series of feature vectors, one containing only single-track features for
a single participant and the other containing single-track features for two
participants along with the track-pair features.
A training set of between 16 and 200 videos was selected manually from C-D1
for each of these 96 HMMs as positive examples depicting each of the 48 action
classes.
A given video could potentially be included in the training sets for both the
one-track and two-track HMMs for the same action class and even for HMMs for
different action classes, if the video was deemed to depict both action
classes.

During testing, we generate present/absent judgments for each video in the test
set paired with each of the 48 action classes.
We do this by thresholding the likelihoods produced by the HMMs.
By varying these thresholds, we can produce an ROC curve for each action class,
comparing the resulting machine-generated present/absent judgments with the
Amazon Mechanical Turk judgments.
When doing so, we test videos for which our tracker produces two or more tracks
against only the two-track HMMs while we test ones for which our
tracker produces a single track against only the one-track HMMs.

\begin{figure*}
  \begin{center}
    \begin{tabular}{ccc}
      \includegraphics[width=0.9\textwidth]{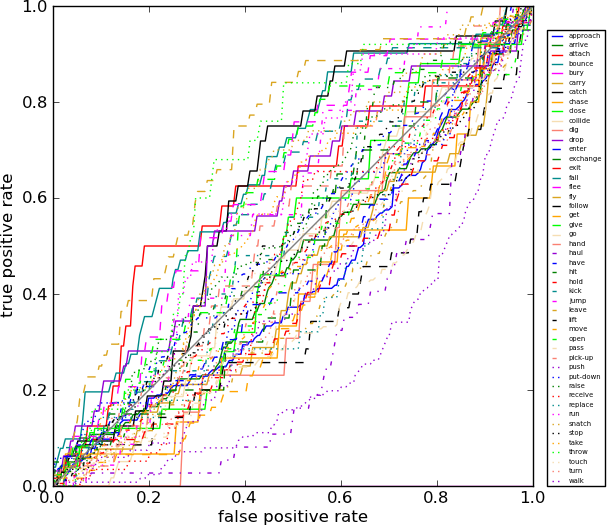}\\
    \end{tabular}
  \end{center}
  \caption{ROC curves for each of the 48 action classes for Experiment~I
    omitting all discrete and body-posture-related features.}
  \label{fig:ROC1}
\end{figure*}

\begin{figure*}
  \begin{center}
    \begin{tabular}{ccc}
      \includegraphics[width=0.9\textwidth]{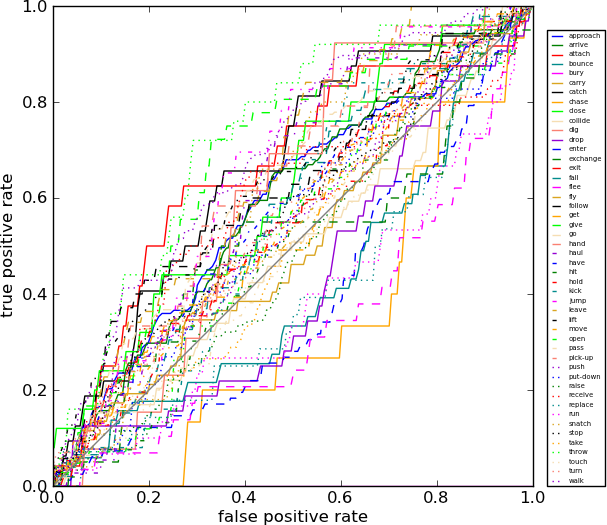}\\
    \end{tabular}
  \end{center}
  \caption{ROC curves for each of the 48 action classes for Experiment~II
    omitting only the discrete features.}
  \label{fig:ROC2}
\end{figure*}

\begin{figure*}
  \begin{center}
    \begin{tabular}{ccc}
      \includegraphics[width=0.9\textwidth]{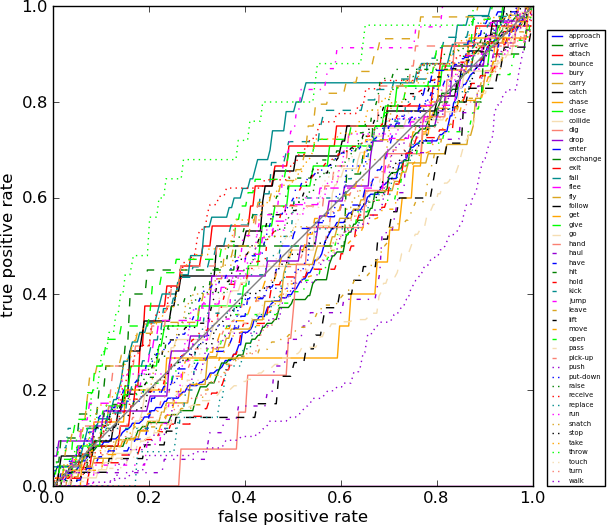}
    \end{tabular}
  \end{center}
  \caption{ROC curves for each of the 48 action classes for Experiment~III
    omitting only the continuous body-posture-related.}
  \label{fig:ROC3}
\end{figure*}

We performed three experiments, training 96 different 200-state HMMs for each.
Experiment~I omitted all discrete features and all body-posture related
features.
Experiment~II omitted only the discrete features.
Experiment~III omitted only the continuous body-posture related features.
ROC curves for each experiment are shown in Fig.~\ref{fig:ROC1},
Fig.~\ref{fig:ROC2} and Fig.~\ref{fig:ROC3}.
Note that the incorporation of body-posture information, either in the form of
continuous normalized part displacements or discrete codebook indices, improves
event-recognition accuracy, despite the fact that the part displacements
produced by the Felzenszwalb \etal\ detectors are noisy and appear
unintelligible to the human eye.

\subsection{Generating sentences}
\label{sec:description}

We produce a sentence from a detected action class together with the associated
tracks using the templates from Table~\ref{tab:templates}.
In these templates, words in \emph{italics} denote fixed strings, words in
\textbf{bold} indicate the action class, X and Y denote subject and
object noun phrases, and the categories Adv, $\textrm{PP}_{\textrm{endo}}$, and
$\textrm{PP}_{\textrm{exo}}$ denote adverbs and prepositional-phrase adjuncts to
describe the subject motion.
The processes for generating these noun phrases, adverbs, and
prepositional-phrase adjuncts are described below.
One-track HMMs take that track to be the agent and thus the subject.
For two-track HMMs we choose the mapping from tracks to roles that yields the
higher likelihood and take the agent track to be the subject and the patient
track to be the object except when the action class is either
\textbf{approached} or \textbf{fled}, the agent is (mostly)
stationary, and the patient moves more than the agent.

\begin{table*}
  \begin{center}
    \scalebox{0.8}
	     {\begin{tabular}{llll}
		 X $[$Adv$]$ \textbf{approached} Y $[\textrm{PP}_{\textrm{exo}}]$&
		 X $[$Adv$]$ \textbf{entered} Y $[\textrm{PP}_{\textrm{endo}}]$&
		 X \textbf{had} Y&
		 X \textbf{put} Y \textbf{down}\\
		 X \textbf{arrived} $[$Adv$]$ $[\textrm{PP}_{\textrm{exo}}]$&
		 X $[$Adv$]$ \textbf{exchanged} \emph{an} \emph{object} \emph{with} Y&
		 X \textbf{hit} $[$\emph{something} \emph{with}$]$ Y&
		 X \textbf{raised} Y\\
		 X $[$Adv$]$ \textbf{attached} \emph{an} \emph{object} \emph{to} Y&
		 X $[$Adv$]$ \textbf{exited} Y $[\textrm{PP}_{\textrm{endo}}]$&
		 X \textbf{held} Y&
		 X \textbf{received} $[$\emph{an} \emph{object} \emph{from}$]$Y\\
		 X \textbf{bounced} $[$Adv$]$ $[\textrm{PP}_{\textrm{endo}}]$&
		 X \textbf{fell} $[$Adv$]$ $[$\emph{because} \emph{of} Y$]$ $[\textrm{PP}_{\textrm{endo}}]$&
		 X \textbf{jumped} $[$Adv$]$ $[$\emph{over} Y$]$ $[\textrm{PP}_{\textrm{endo}}]$&
		 X $[$Adv$]$ \textbf{replaced} Y\\
		 X \textbf{buried} Y&
		 X \textbf{fled} $[$Adv$]$ $[$\emph{from} Y$]$ $[\textrm{PP}_{\textrm{endo}}]$&
		 X $[$Adv$]$ \textbf{kicked} Y $[\textrm{PP}_{\textrm{endo}}]$&
		 X \textbf{ran} $[$Adv$]$ $[$\emph{to} Y$]$ $[\textrm{PP}_{\textrm{endo}}]$\\
		 X $[$Adv$]$ \textbf{carried} Y $[\textrm{PP}_{\textrm{endo}}]$&
		 X \textbf{flew} $[$Adv$]$ $[\textrm{PP}_{\textrm{endo}}]$&
		 X \textbf{left} $[$Adv$]$ $[\textrm{PP}_{\textrm{endo}}]$&
		 X $[$Adv$]$ \textbf{snatched} \emph{an} \emph{object} \emph{from} Y\\
		 X \textbf{caught} Y $[\textrm{PP}_{\textrm{exo}}]$&
		 X $[$Adv$]$ \textbf{followed} Y $[\textrm{PP}_{\textrm{endo}}]$&
		 X $[$Adv$]$ \textbf{lifted} Y&
		 X $[$Adv$]$ \textbf{stopped} $[$Y$]$\\
		 X $[$Adv$]$ \textbf{chased} Y $[\textrm{PP}_{\textrm{endo}}]$&
		 X \textbf{got} \emph{an} \emph{object} \emph{from} Y&
		 X $[$Adv$]$ \textbf{moved} Y $[\textrm{PP}_{\textrm{endo}}]$&
		 X $[$Adv$]$ \textbf{took} \emph{an} \emph{object} \emph{from} Y\\
		 X \textbf{closed} Y&
		 X \textbf{gave} \emph{an} \emph{object} \emph{to} Y&
		 X \textbf{opened} Y&
		 X $[$Adv$]$ \textbf{threw} Y $[\textrm{PP}_{\textrm{endo}}]$\\
		 X $[$Adv$]$ \textbf{collided} \emph{with} Y $[\textrm{PP}_{\textrm{exo}}]$&
		 X \textbf{went} $[$Adv$]$ \emph{away} $[\textrm{PP}_{\textrm{endo}}]$&
		 X $[$Adv$]$ \textbf{passed} Y $[\textrm{PP}_{\textrm{exo}}]$&
		 X \textbf{touched} Y\\
		 X \emph{was} \textbf{digging} $[$\emph{with} Y$]$&
		 X \textbf{handed} Y \emph{an} \emph{object}&
		 X \textbf{picked} Y \textbf{up}&
		 X \textbf{turned} $[\textrm{PP}_{\textrm{endo}}]$\\
		 X \textbf{dropped} Y&
		 X $[$Adv$]$ \textbf{hauled} Y $[\textrm{PP}_{\textrm{endo}}]$&
		 X $[$Adv$]$ \textbf{pushed} Y $[\textrm{PP}_{\textrm{endo}}]$&
		 X \textbf{walked} $[$Adv$]$ $[$\emph{to} Y$]$ $[\textrm{PP}_{\textrm{endo}}]$
	      \end{tabular}}
  \end{center}
  \caption{Sentential templates for the action classes indicated in bold.}
  \label{tab:templates}
\end{table*}

Brackets in the templates denote optional entities.
Optional entities containing Y are generated only for two-track HMMs.
The criteria for generating optional adverbs and prepositional phrases are
described below.
The optional entity for \textbf{received} is generated when there is a
patient track whose category is \texttt{mailbox}, \texttt{person},
\texttt{person-crouch}, or \texttt{person-down}.

We use adverbs to describe the velocity of the subject.
For some verbs, a velocity adverb would be awkward:
\begin{small}
  \begin{center}
    \begin{tabular}{@{}l@{}l@{}}
      \parbox{0.45\columnwidth}{$*$X \textbf{\emph{slowly}} \emph{had} Y}&
      \parbox{0.45\columnwidth}{$*$X \emph{had} \textbf{\emph{slowly}} Y}
    \end{tabular}
  \end{center}
\end{small}
\noindent
Furthermore, stylistic considerations dictate the syntactic position of an
optional adverb:
\begin{small}
  \begin{center}
    \begin{tabular}{@{}l@{}l@{}}
      \parbox{0.45\columnwidth}{X jumped \textbf{\emph{slowly}} over Y}&
      \parbox{0.45\columnwidth}{X \textbf{\emph{slowly}} jumped over Y}\\
      \parbox{0.45\columnwidth}{X \textbf{\emph{slowly}} \emph{approached} Y}&
      \parbox{0.45\columnwidth}{$*$X \emph{approached} \textbf{\emph{slowly}} Y}\\
      \parbox{0.45\columnwidth}{$?$X \textbf{\emph{slowly}} \emph{fell}}&
      \parbox{0.45\columnwidth}{X \emph{fell} \textbf{\emph{slowly}}}\\
    \end{tabular}
  \end{center}
\end{small}
\noindent
The verb-phrase templates thus indicate whether an adverb is allowed, and if so
whether it occurs, preferentially, preverbally or postverbally.
Adverbs are chosen subject to three thresholds~$v^{\textrm{action class}}_1$,
$v^{\textrm{action class}}_2$, and~$v^{\textrm{action class}}_3$ determined
empirically on a per-action-class basis:
We select those frames from the subject track where the magnitude of the
velocity of the box-detection center is above~$v^{\textrm{action class}}_1$.
An optional adverb is generated by comparing the magnitude of the average
velocity~$v$ of the subject track box-detection centers in these frames to the
per-action-class thresholds:
\begin{displaymath}
  \begin{array}{ll}
    \text{\emph{quickly}}&v>v^{\textrm{action class}}_2\\
    \text{\emph{slowly}}&v^{\textrm{action class}}_1\leq v\leq v^{\textrm{action
    class}}_3
  \end{array}
\end{displaymath}

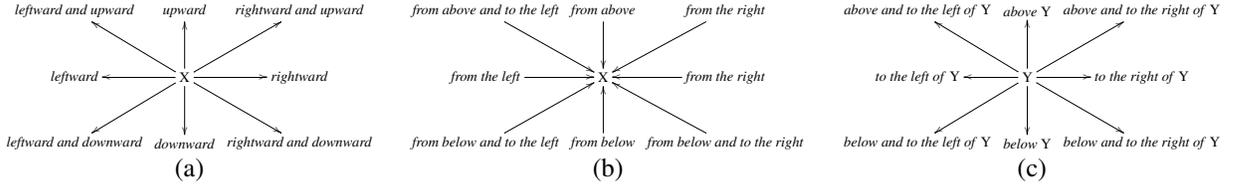
\begin{figure*}
  \vspace{3ex}
  \begin{center}
    \begin{tabular}{@{}ccc@{}}
      \scalebox{0.54}{\begin{math}
	  \xymatrix@R=7ex@C=0.25em{
	    *+{\mbox{\emph{leftward and upward}}}&
	    *+{\mbox{\emph{upward}}}&
	    *+{\mbox{\emph{rightward and upward}}}\\
	    *+{\mbox{\emph{leftward}}}&
	    *+{\mbox{X}}\ar[l]\ar[ul]\ar[dl]\ar[u]\ar[d]\ar[ur]\ar[dr]\ar[r]&
	    *+{\mbox{\emph{rightward}}}\\
	    *+{\mbox{\emph{leftward and downward}}}&
	    *+{\mbox{\emph{downward}}}&
	    *+{\mbox{\emph{rightward and downward}}}}
      \end{math}}&
      \scalebox{0.54}{\begin{math}
	  \xymatrix@R=7ex@C=0.25em{
	    *+{\mbox{\emph{from above and to the left}}}\ar[dr]&
	    *+{\mbox{\emph{from above}}}\ar[d]&
	    *+{\mbox{\emph{from the right}}}\ar[dl]\\
	    *+{\mbox{\emph{from the left}}}\ar[r]&
	    *+{\mbox{X}}&
	    *+{\mbox{\emph{from the right}}}\ar[l]\\
	    *+{\mbox{\emph{from below and to the left}}}\ar[ur]&
	    *+{\mbox{\emph{from below}}}\ar[u]&
	    *+{\mbox{\emph{from below and to the right}}}\ar[ul]}
      \end{math}}&
      \scalebox{0.54}{\begin{math}
	  \xymatrix@R=7ex@C=0.25em{
	  *+{\mbox{\emph{above and to the left of} Y}}&
	  *+{\mbox{\emph{above} Y}}&
	  *+{\mbox{\emph{above and to the right of} Y}}\\
	  *+{\mbox{\emph{to the left of} Y}}&
	  *+{\mbox{Y}}\ar[l]\ar[ul]\ar[dl]\ar[u]\ar[d]\ar[ur]\ar[dr]\ar[r]&
	  *+{\mbox{\emph{to the right of} Y}}\\
	  *+{\mbox{\emph{below and to the left of} Y}}&
	  *+{\mbox{\emph{below} Y}}&
	  *+{\mbox{\emph{below and to the right of} Y}}}
      \end{math}}\\
      (a) & (b) & (c)
    \end{tabular}
  \end{center}
  \caption{(a)~Endogenous and (b)~exogenous prepositional-phrase adjuncts to
    describe subject motion direction.
    (c)~Prepositional phrases incorporated into subject noun phrases
    describing viewer-relative 2D spatial relations between the subject X and
    the reference object Y.\@}
  \label{fig:adjuncts-and-prepositional-phrases}
\end{figure*}

We use prepositional-phrase adjuncts to describe the motion direction of the
subject.
Again, for some verbs, such adjuncts would be awkward:
\begin{small}
  \begin{center}
    \begin{tabular}{@{}l@{}l@{}}
      \parbox{0.45\columnwidth}{$*$X \emph{had} Y \textbf{\emph{leftward}}}&
      \parbox{0.45\columnwidth}{$*$X \emph{had} Y \textbf{\emph{from} \emph{the} \emph{left}}}
    \end{tabular}
  \end{center}
\end{small}
\noindent
Moreover, for some verbs it is natural to describe the motion direction
endogenously, from the perspective of the subject, while for others it is more
natural to describe the motion direction exogenously, from the perspective of
the viewer:
\begin{small}
  \begin{center}
    \begin{tabular}{@{}l@{}l@{}}
      \parbox{0.45\columnwidth}{X \emph{fell} \textbf{\emph{leftward}}}&
      \parbox{0.45\columnwidth}{X \emph{fell} \textbf{\emph{from} \emph{the} \emph{left}}}\\
      \parbox{0.45\columnwidth}{X \emph{chased} Y \textbf{\emph{leftward}}}&
      \parbox{0.45\columnwidth}{$*$X \emph{chased} Y \textbf{\emph{from} \emph{the} \emph{left}}}\\
      \parbox{0.45\columnwidth}{$*$X \emph{arrived} \textbf{\emph{leftward}}}&
      \parbox{0.45\columnwidth}{X \emph{arrived} \textbf{\emph{from} \emph{the} \emph{left}}}\\
    \end{tabular}
  \end{center}
\end{small}
\noindent
The verb-phrase templates thus indicate whether an adjunct is allowed, and
if so whether it is preferentially endogenous or exogenous.
The choice of adjunct is determined from the orientation of~$v$, as computed
above and depicted in Fig.~\ref{fig:adjuncts-and-prepositional-phrases}(a,b).
We omit the adjunct when $v<v^{\textrm{action class}}_1$.

We generate noun phrases X and Y to refer to event participants according to
the following grammar:
\begin{displaymath}
  \begin{array}{@{\hspace{0ex}}l@{\hspace{2ex}}c@{\hspace{2ex}}l@{\hspace{0ex}}}
    \textrm{NP}&\rightarrow&\text{\emph{themselves}}\;|\;
    \text{\emph{itself}}\;|\;\text{\emph{something}}\;|\;
    \textrm{D}\;\textrm{A}^{*}\;\textrm{N}\;[\textrm{PP}]\\
    \textrm{D}&\rightarrow&\text{\emph{the}}\;|\;\text{\emph{that}}\;|\;
    \text{\emph{some}}
  \end{array}
\end{displaymath}
\noindent
When instantiating a sentential template that has a required object noun-phrase
Y for a one-track HMM, we generate a pronoun.
A pronoun is also generated when the action class is
\textbf{entered} or \textbf{exited} and the patient class is not
\texttt{car}, \texttt{door}, \texttt{suv}, or \texttt{truck}.
The anaphor \emph{themselves} is generated if the action class is
\textbf{attached} or \textbf{raised}, the anaphor \emph{itself}
if the action class is \textbf{moved}, and \emph{something} otherwise.

As described below, we generate an optional prepositional phrase for the
subject noun phrase to describe the spatial relation between the subject and
the object.
We choose the determiner to handle coreference, generating \emph{the} when a
noun phrase unambiguously refers to the agent or the patient due to the
combination of head noun and any adjectives,
\begin{small}
  \begin{quote}
    \emph{\textbf{The} person jumped over \textbf{the} ball.}\\
    \emph{\textbf{The} red ball collided with \textbf{the} blue ball.}
  \end{quote}
\end{small}
\emph{that} for an object noun phrase that corefers to a track referred to in a
prepositional phrase for the subject,
\begin{small}
  \begin{quote}
    \emph{\textbf{The} person to the right of \textbf{the} car approached
      \textbf{that} car.}\\
    \emph{\textbf{Some} person to the right of \textbf{some} other person
      approached \textbf{that} other person.}
  \end{quote}
\end{small}
and \emph{some} otherwise:
\begin{small}
  \begin{quote}
    \emph{\textbf{Some} car approached \textbf{some} other car.}
  \end{quote}
\end{small}

We generate the head noun of a noun phrase from the object class using the
mapping in Table~\ref{tab:objects}(a).
Four different kinds of adjectives are generated: color, shape, size, and
restrictive modifiers.
An optional color adjective is generated based on the average HSV values in the
eroded detection boxes for a track: \emph{black} when $V\leq0.2$, \emph{white}
when $V\geq0.8$, one of \emph{red}, \emph{blue}, \emph{green}, \emph{yellow},
\emph{teal}, or \emph{pink} based on $H$, when $S\geq0.7$.
An optional size adjective is generated in two ways, one from the object class
using the mapping in Table~\ref{tab:objects}(c), the other based on
per-object-class image statistics.
For each object class, a mean object size $\bar{a}_{\textrm{object class}}$ is
determined by averaging the detected-box areas over all tracks for that object
class in the training set used to train HMMs.
An optional size adjective for a track is generated by comparing the average
detected-box area~$a$ for that track to $\bar{a}_{\textrm{object class}}$:
\begin{displaymath}
  \begin{array}{ll}
    \text{\emph{big}}&a\geq\beta_{\textrm{object class}}\bar{a}_{\textrm{object class}}\\
    \text{\emph{small}}&a\leq\alpha_{\textrm{object class}}\bar{a}_{\textrm{object class}}
  \end{array}
\end{displaymath}
\noindent
The per-object-class cutoff ratios $\alpha_{\textrm{object class}}$ and
$\beta_{\textrm{object class}}$ are computed to equally tripartition the
distribution of per-object-class mean object sizes on the training set.
Optional shape adjectives are generated in a similar fashion.
Per-object-class mean aspect ratios $\bar{r}_{\text{object class}}$ are
determined in addition to the per-object-class mean object sizes
$\bar{a}_{\text{object class}}$.
Optional shape adjectives for a track are generated by comparing the average
detected-box aspect ratio~$r$ and area~$a$ for that track to these means:
\begin{displaymath}
  \begin{array}{@{\hspace{0ex}}ll@{\hspace{0ex}}}
    \text{\emph{tall}}&r\leq0.7\bar{r}_{\text{object class}}\wedge
                       a\geq\beta_{\textrm{\tiny object class}}\bar{a}_{\textrm{\tiny object class}}\\
    \text{\emph{short}}&r\geq1.3\bar{r}_{\text{object class}}\wedge
                        a\leq\alpha_{\textrm{\tiny object class}}\bar{a}_{\textrm{\tiny object class}}\\
    \text{\emph{narrow}}&r\leq0.7\bar{r}_{\text{object class}}\wedge
                         a\leq\alpha_{\textrm{\tiny object class}}\bar{a}_{\textrm{\tiny object class}}\\
    \text{\emph{wide}}&r\geq1.3\bar{r}_{\text{object class}}\wedge
                       a\geq\beta_{\textrm{\tiny object class}}\bar{a}_{\textrm{\tiny object class}}\\
  \end{array}
\end{displaymath}
\noindent
To avoid generating shape and size adjectives for unstable tracks, they are
only generated when the detection-score variance and the detected aspect-ratio
variance for the track are below specified thresholds.
Optional restrictive modifiers are generated from the object class using the
mapping in Table~\ref{tab:objects}(b).
Person-pose adjectives are generated from aggregate body-posture information
for the track: object class, normalized part displacements, and body-posture
codebook indices.
We generate all applicable adjectives except for color and person pose.
Following the Gricean Maxim of Quantity \citep{Grice1975}, we only generate
color and person-pose adjectives if needed to prevent coreference of
nonhuman event participants.
Finally, we generate an initial adjective \emph{other}, as needed to prevent
coreference.
Generating \emph{other} does not allow generation of the determiner \emph{the}
in place of \emph{that} or \emph{some}.
We order any adjectives generated so that \emph{other} comes first, followed by
size, shape, color, and restrictive modifiers, in that order.

For two-track HMMs where neither participant moves, a prepositional phrase is
generated for subject noun phrases to describe the static 2D spatial relation
between the subject X and the reference object Y from the perspective
of the viewer, as shown in Fig.~\ref{fig:adjuncts-and-prepositional-phrases}(c).

\section{Experimental results}

\begin{figure*}
  \begin{center}
    \begin{tabular}{@{}c@{\hspace*{2pt}}c@{\hspace*{2pt}}c@{\hspace*{2pt}}c@{}}
      \includegraphics[width=0.24\textwidth]{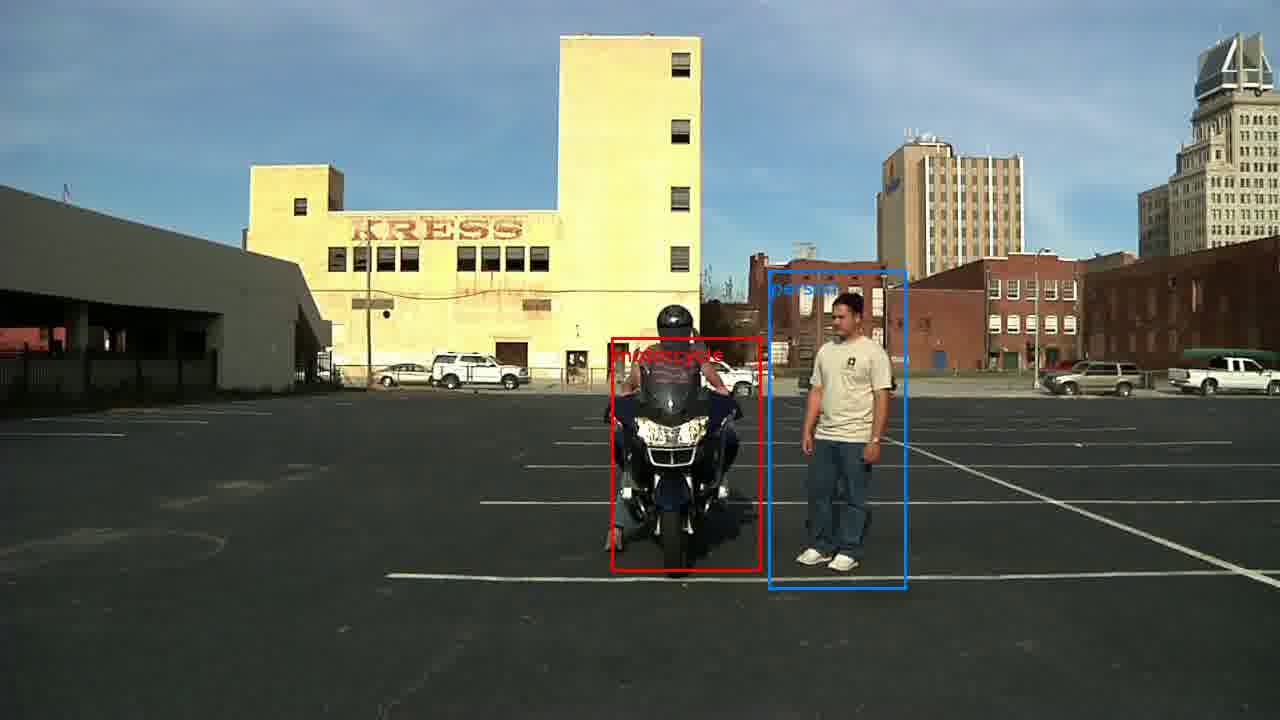}&
      \includegraphics[width=0.24\textwidth]{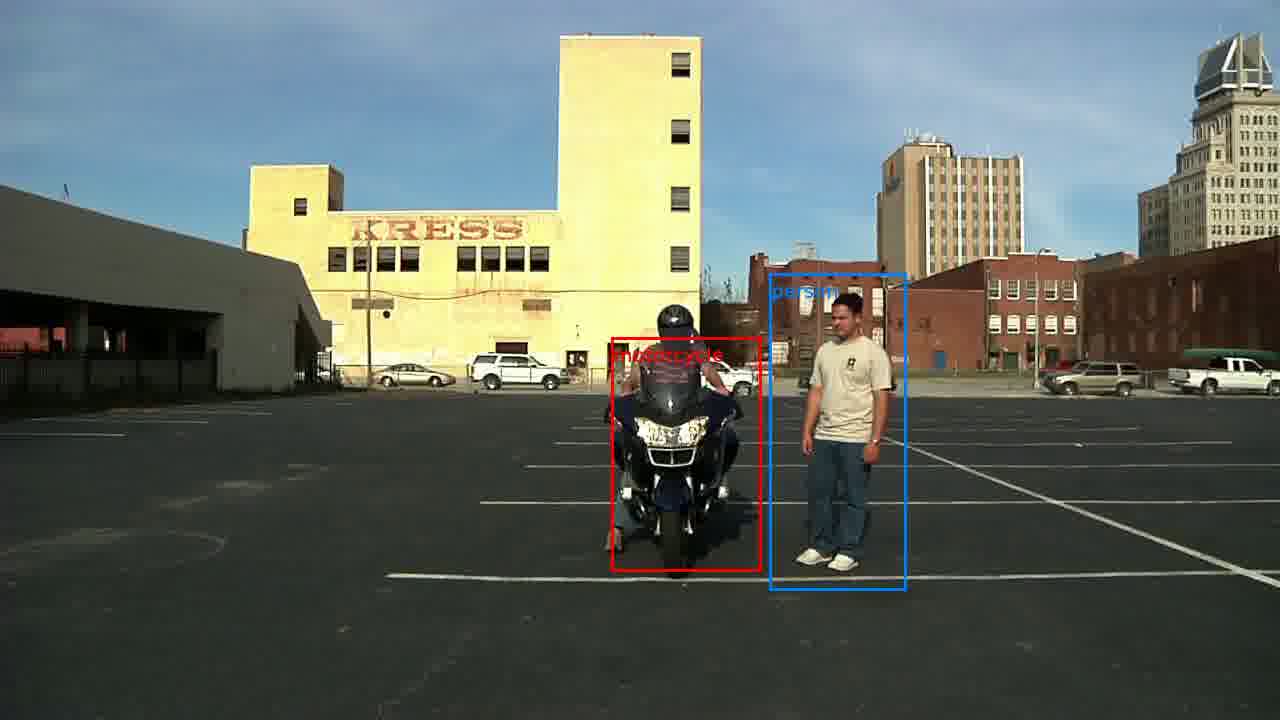}&
      \includegraphics[width=0.24\textwidth]{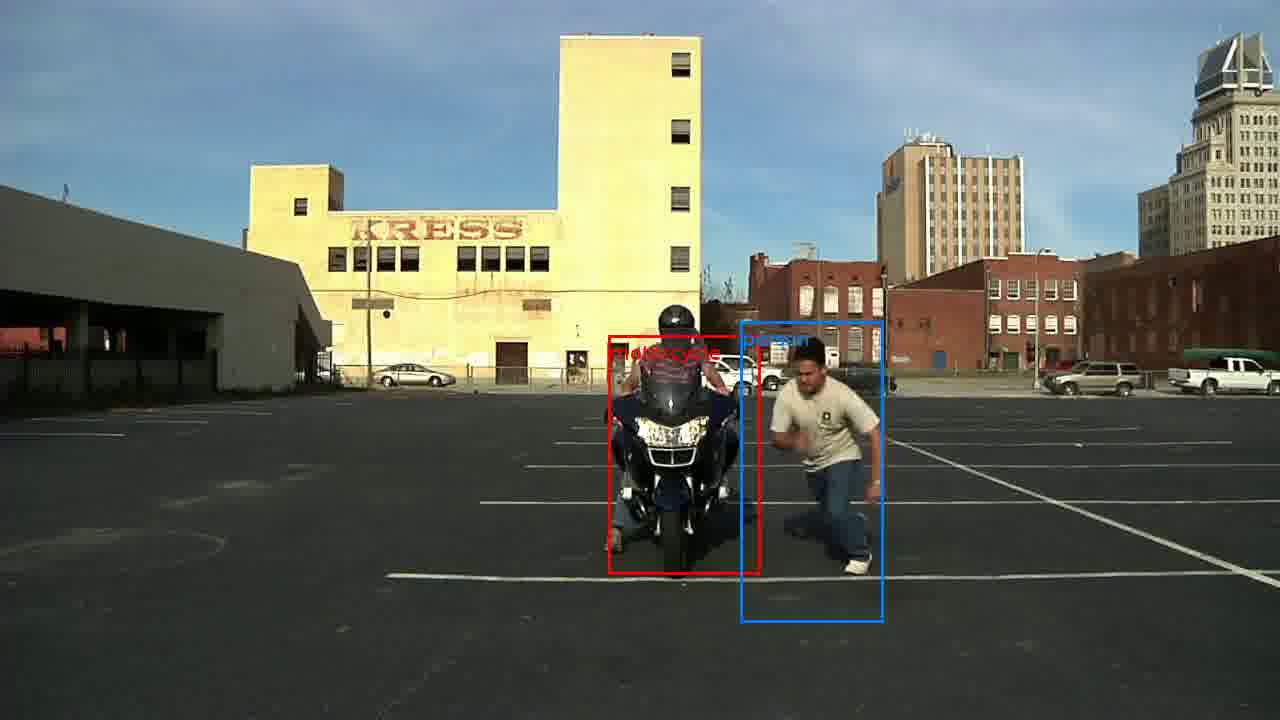}&
      \includegraphics[width=0.24\textwidth]{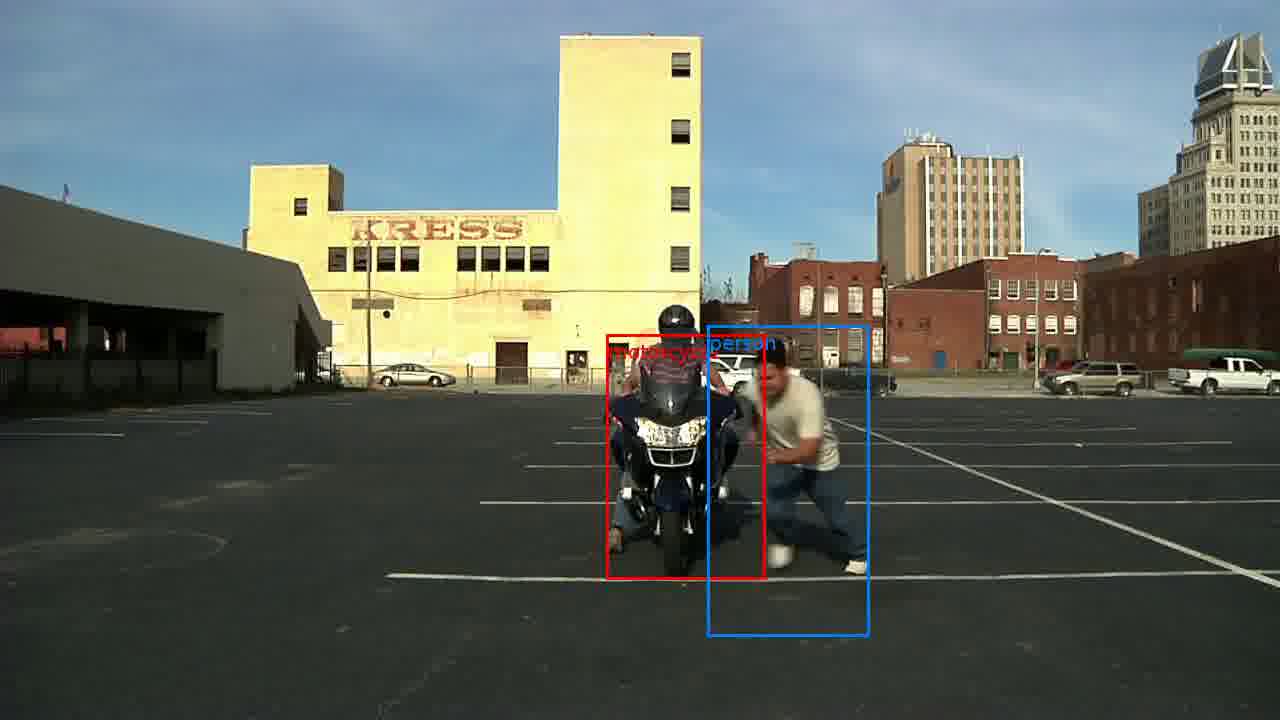}\\
      \includegraphics[width=0.24\textwidth]{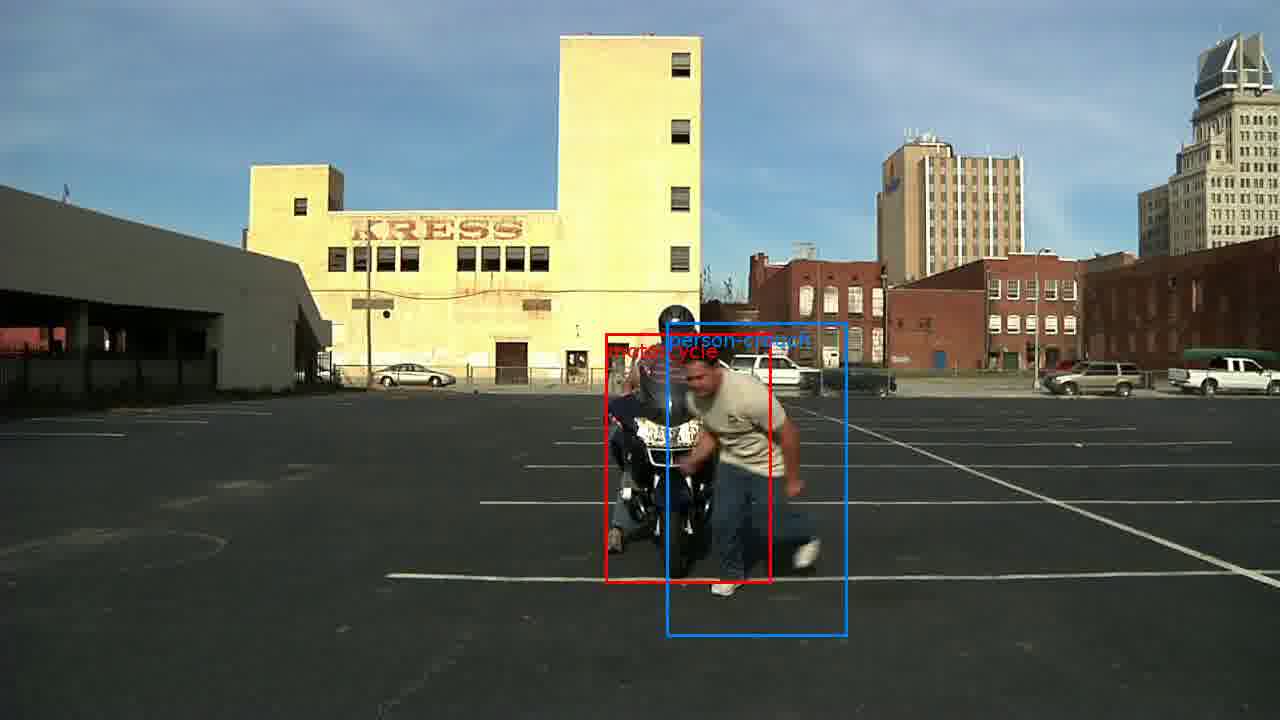}&
      \includegraphics[width=0.24\textwidth]{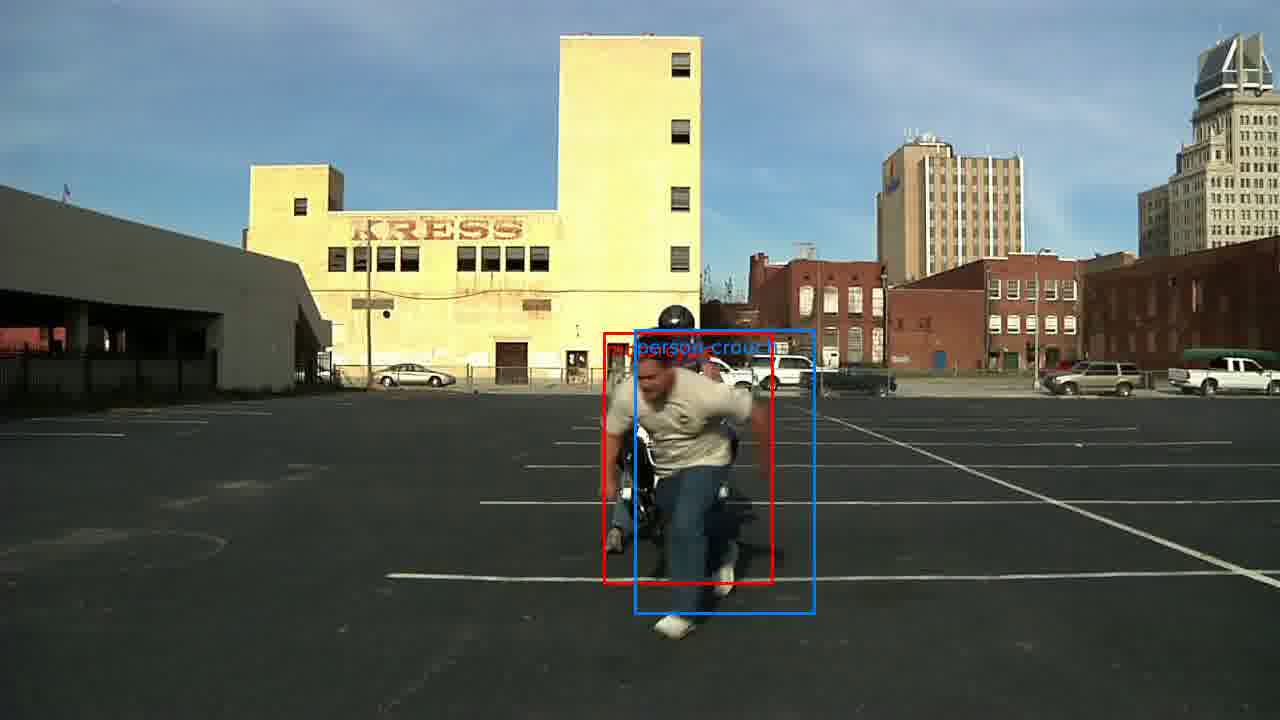}&
      \includegraphics[width=0.24\textwidth]{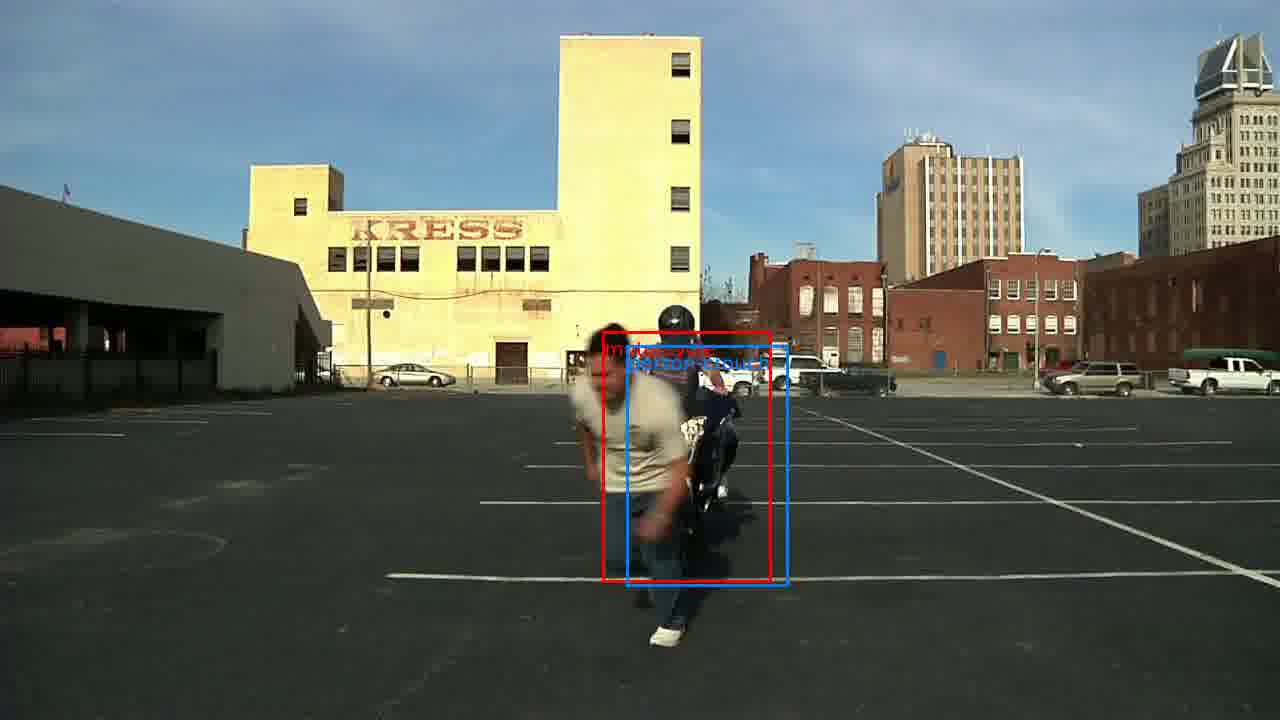}&
      \includegraphics[width=0.24\textwidth]{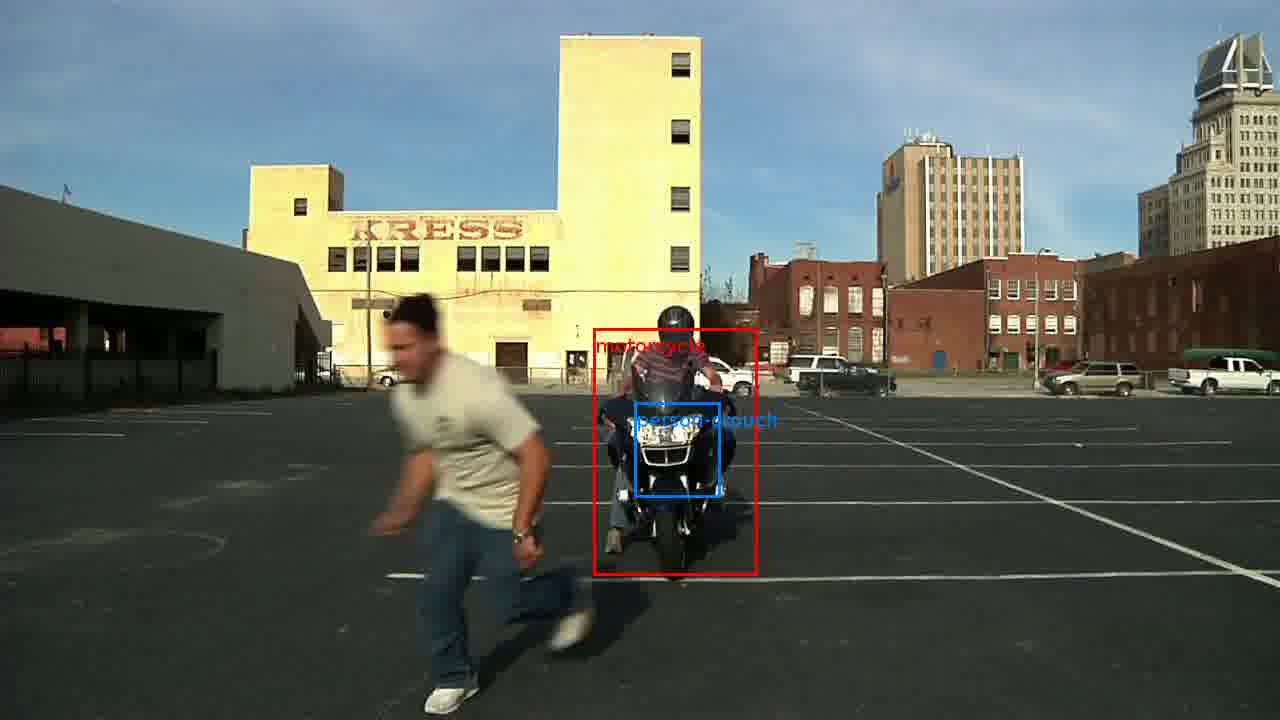}\\
      \multicolumn{4}{c}{The upright person to the right of the motorcycle
        went away leftward.}\\
      \includegraphics[width=0.24\textwidth]{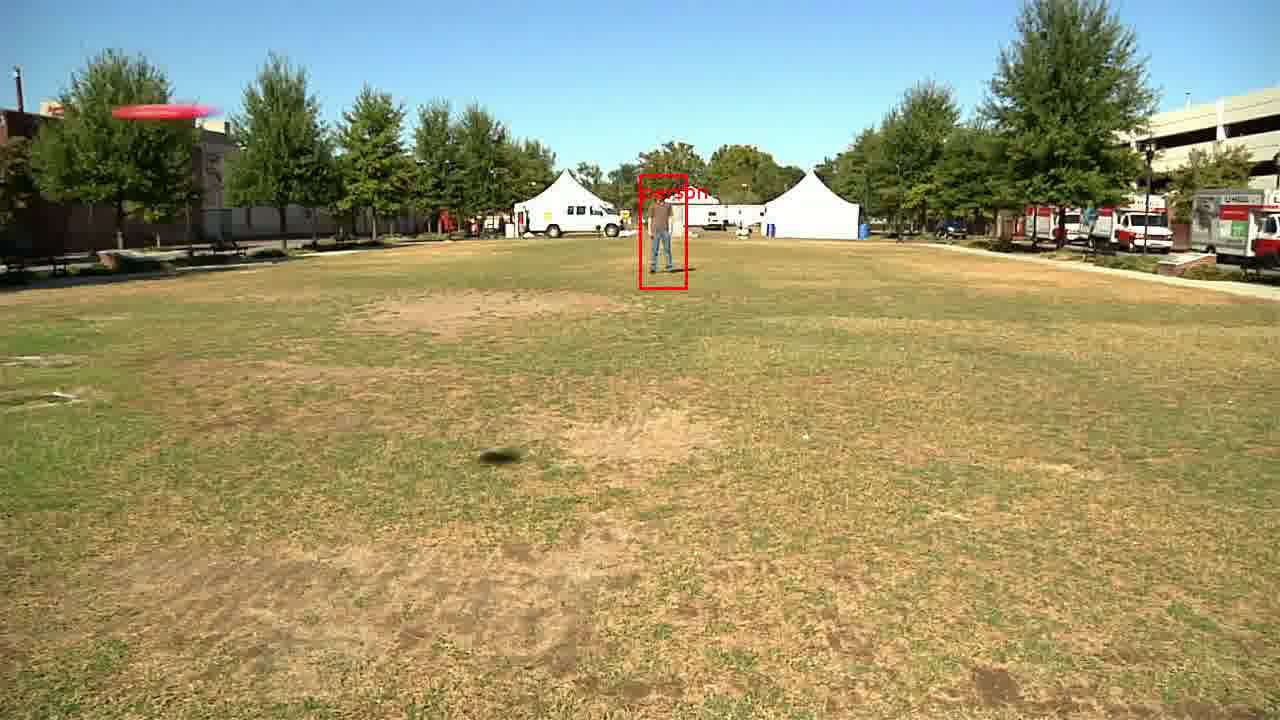}&
      \includegraphics[width=0.24\textwidth]{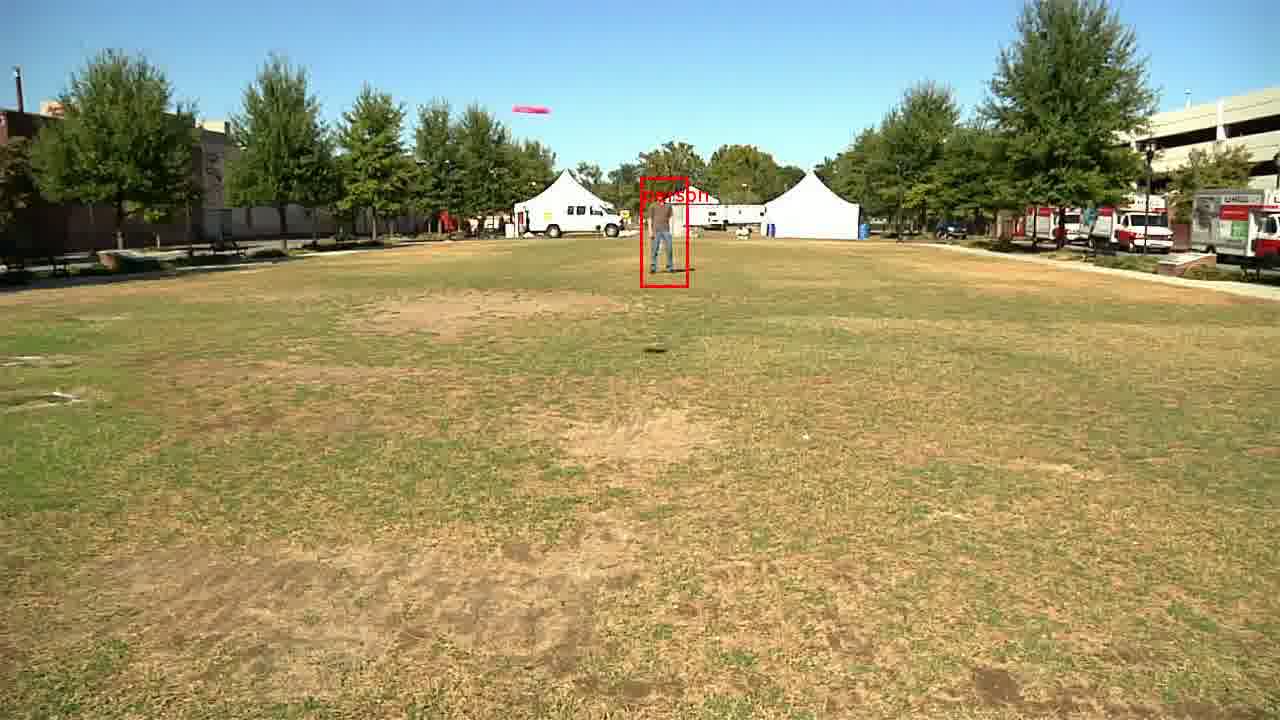}&
      \includegraphics[width=0.24\textwidth]{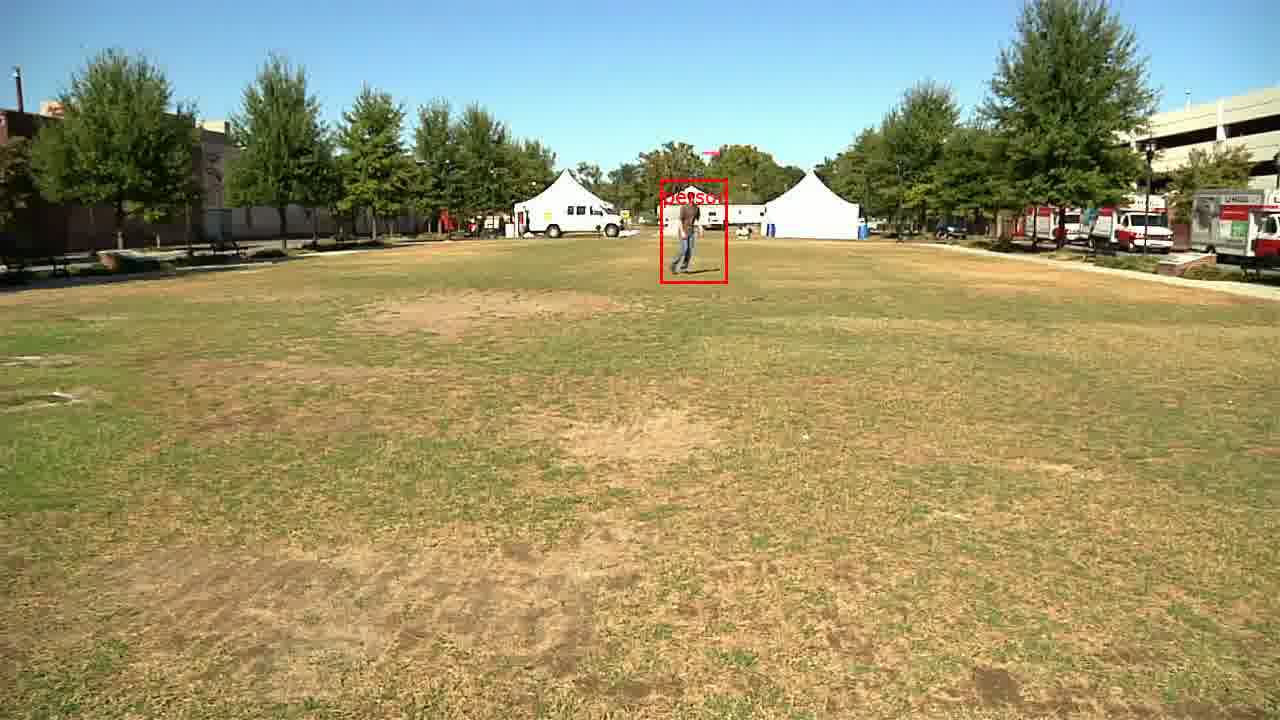}&
      \includegraphics[width=0.24\textwidth]{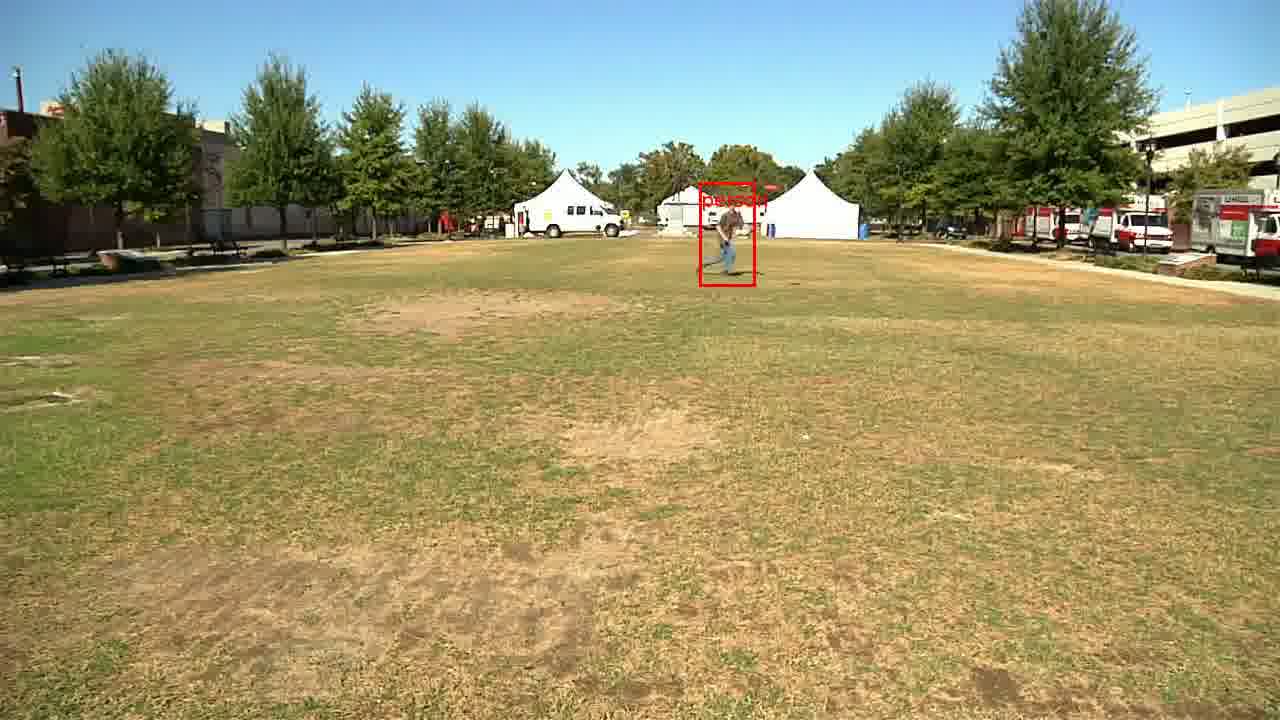}\\
      \includegraphics[width=0.24\textwidth]{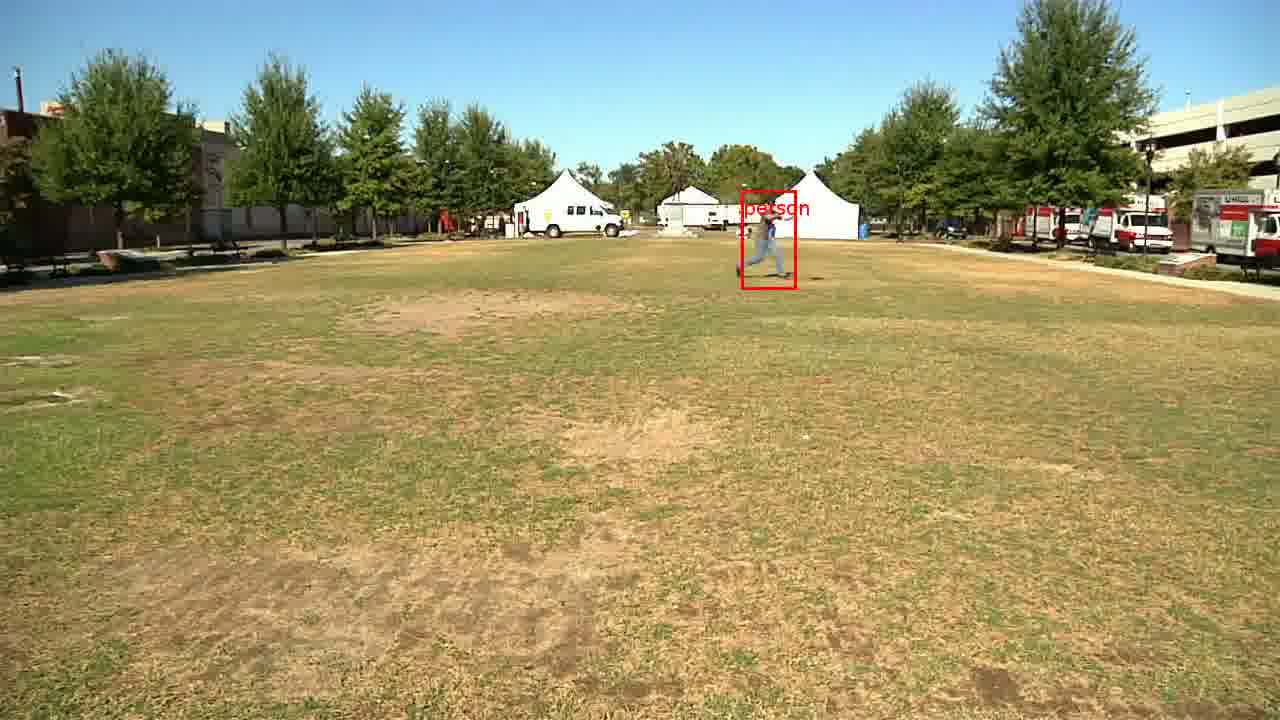}&
      \includegraphics[width=0.24\textwidth]{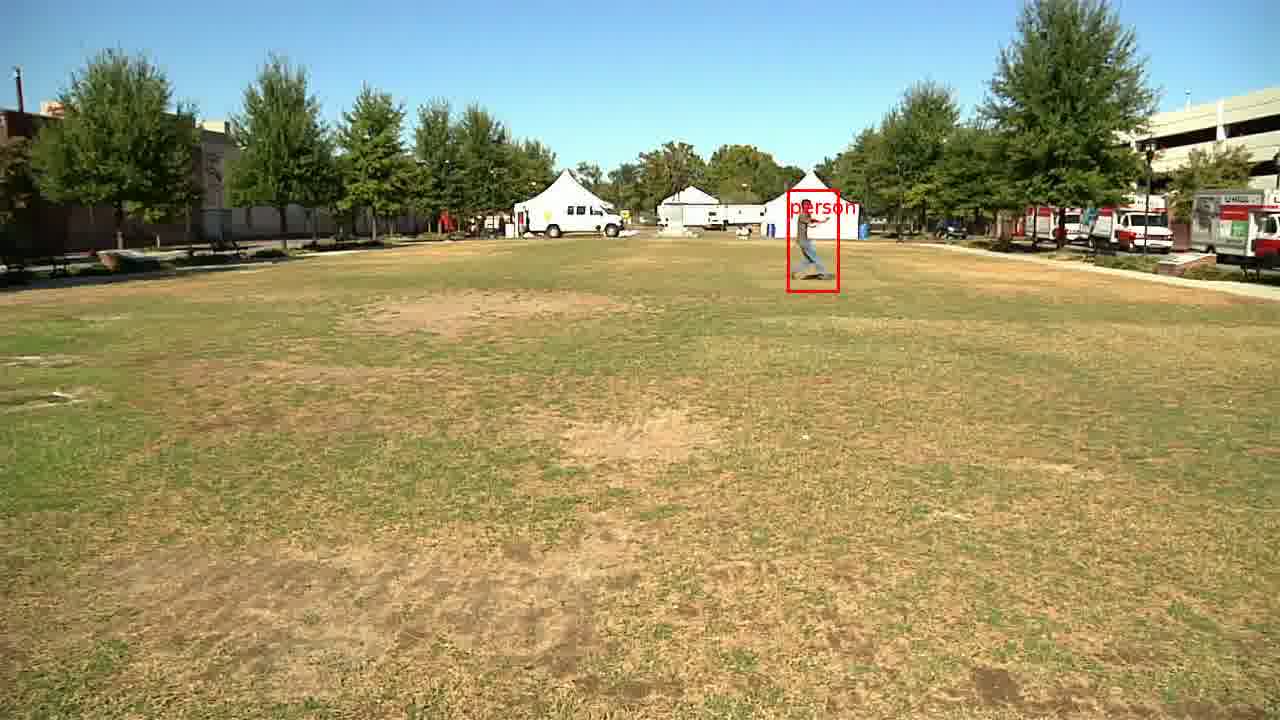}&
      \includegraphics[width=0.24\textwidth]{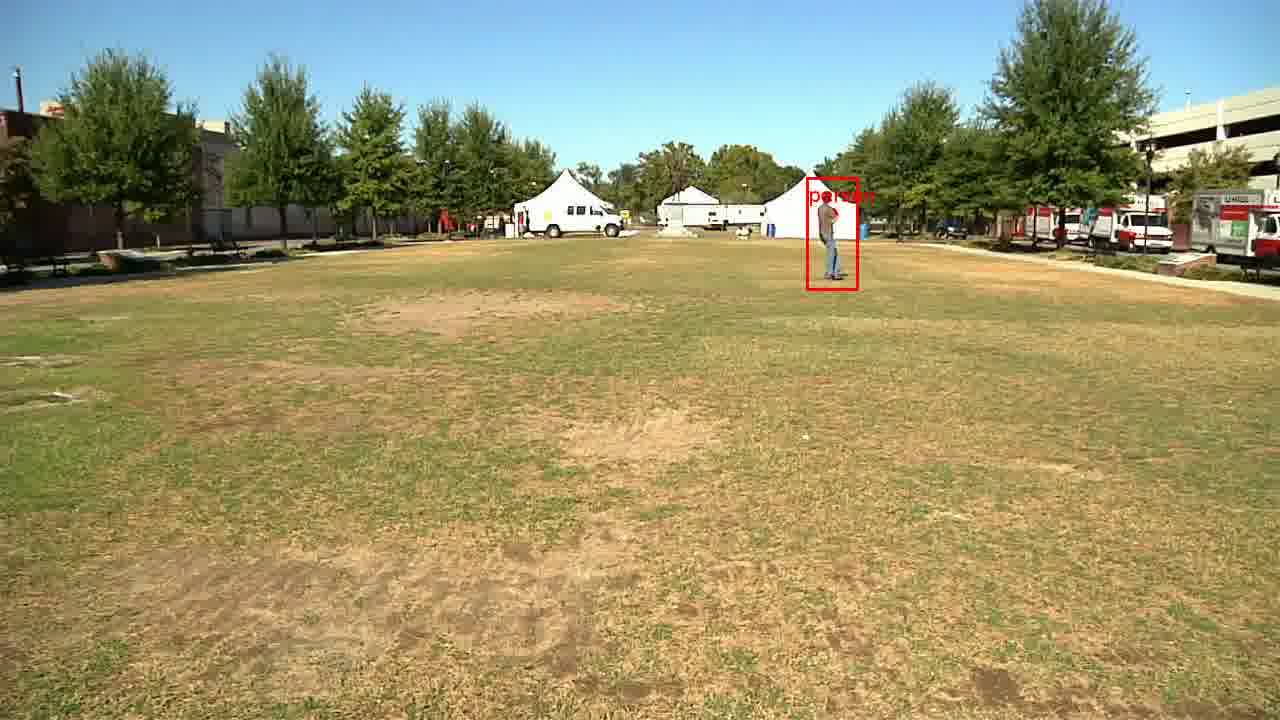}&
      \includegraphics[width=0.24\textwidth]{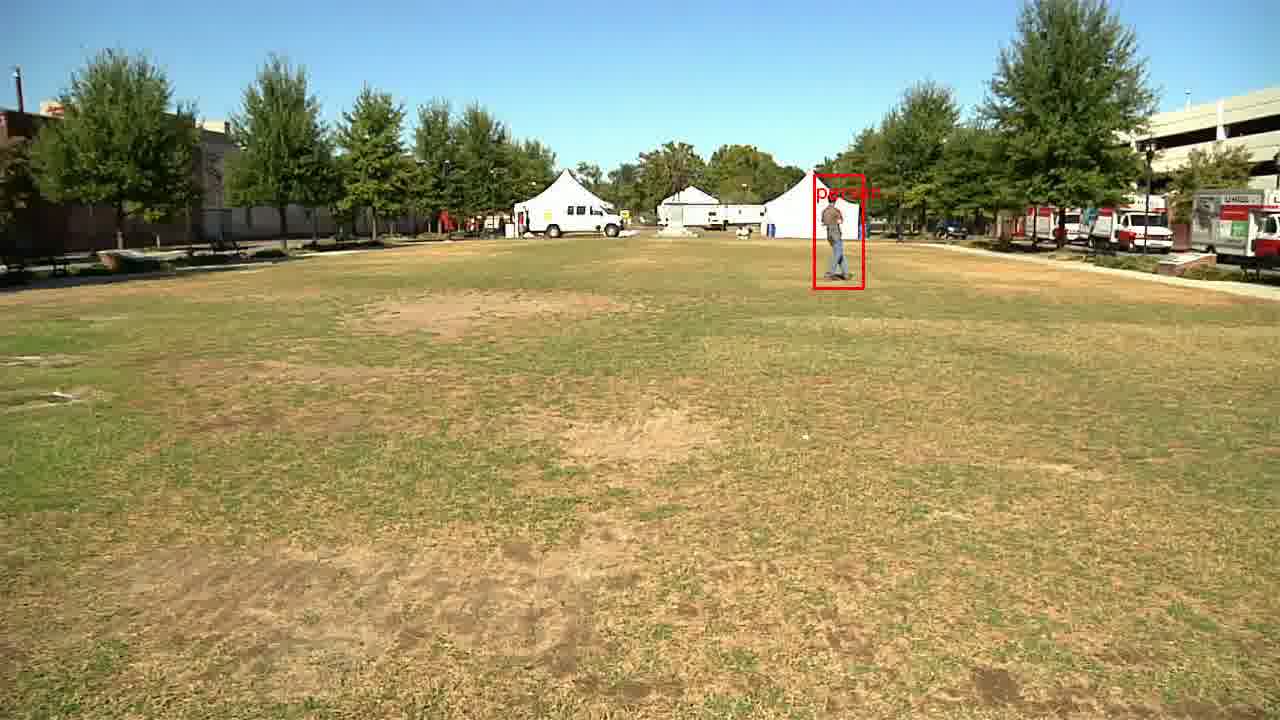}\\
      \multicolumn{4}{c}{The person walked slowly to something rightward.}\\
      \includegraphics[width=0.24\textwidth]{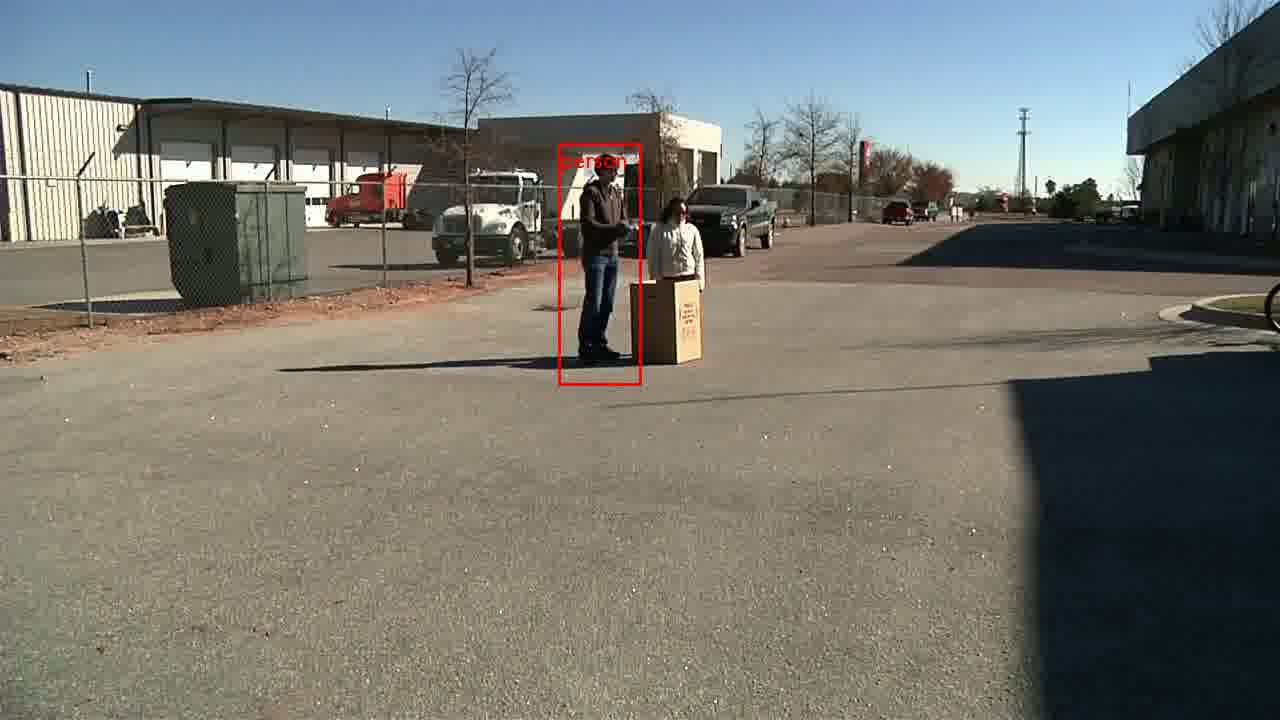}&
      \includegraphics[width=0.24\textwidth]{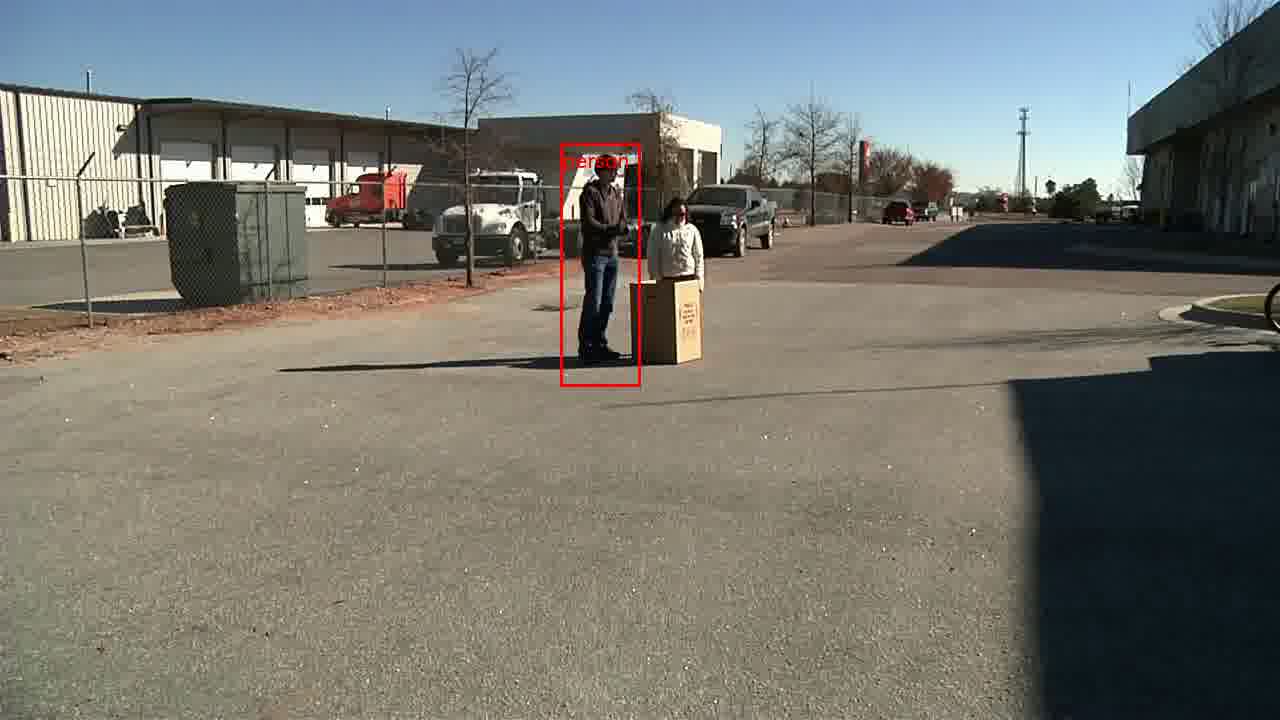}&
      \includegraphics[width=0.24\textwidth]{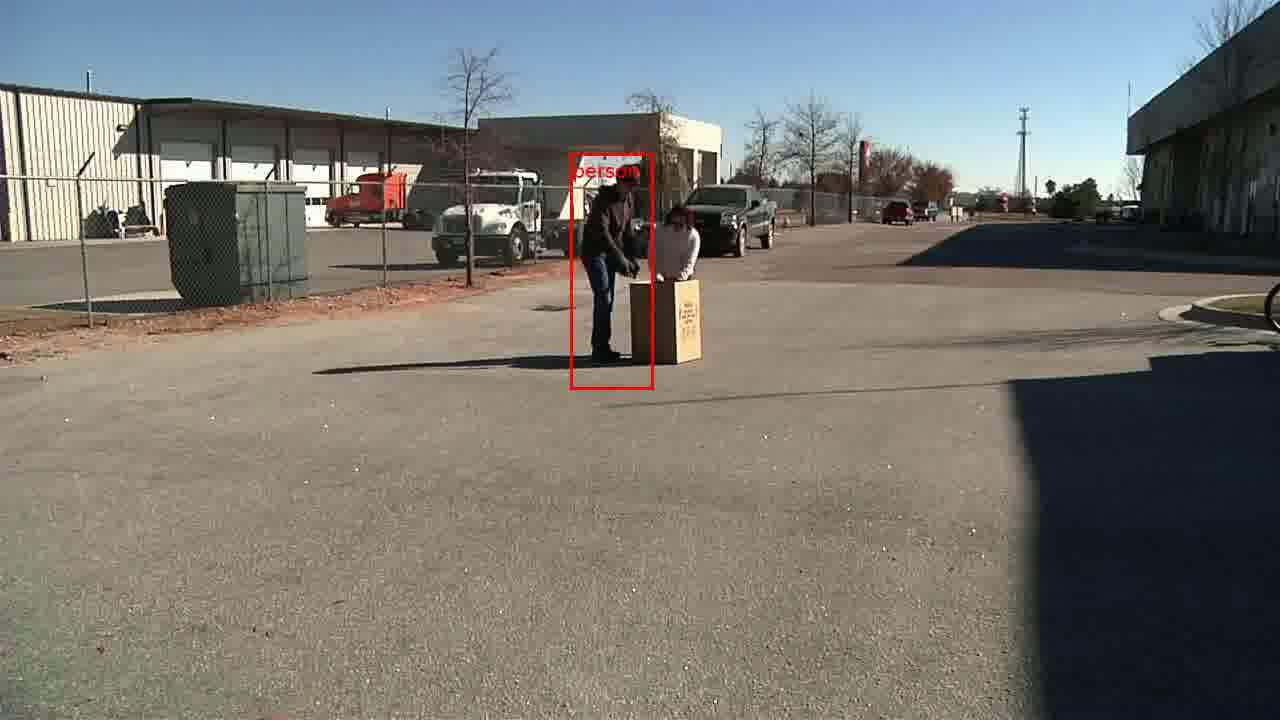}&
      \includegraphics[width=0.24\textwidth]{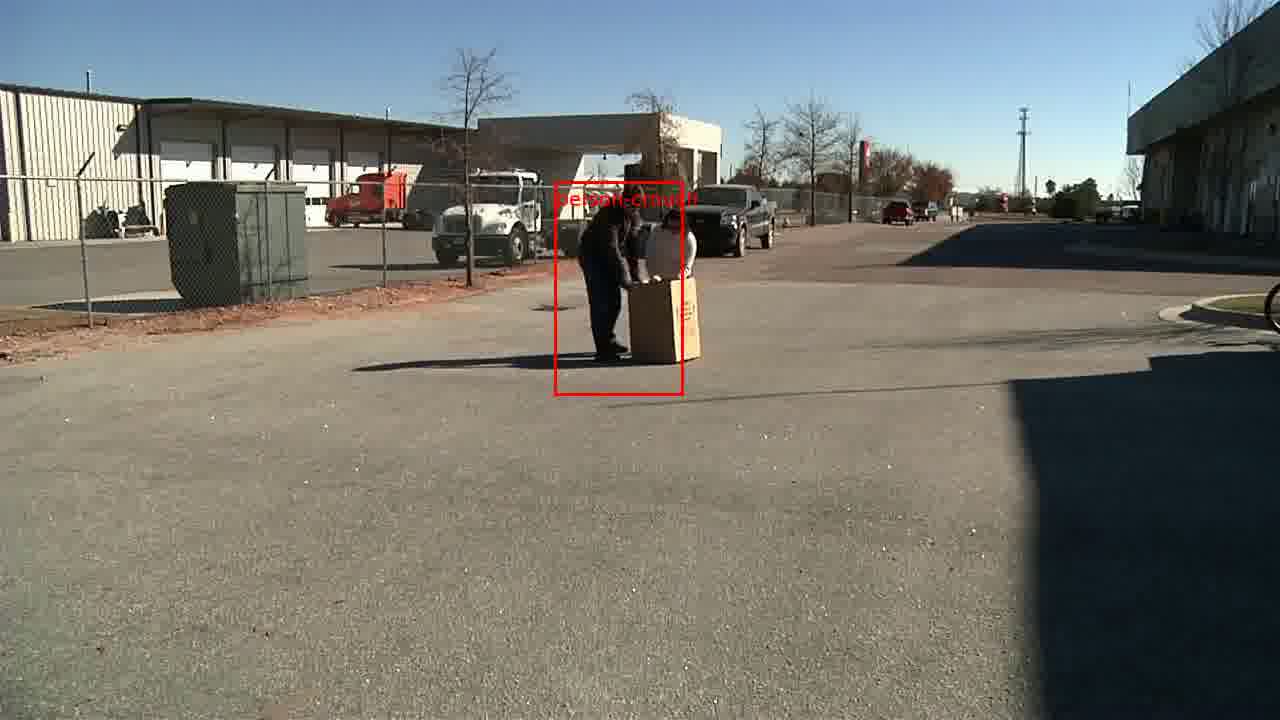}\\
      \includegraphics[width=0.24\textwidth]{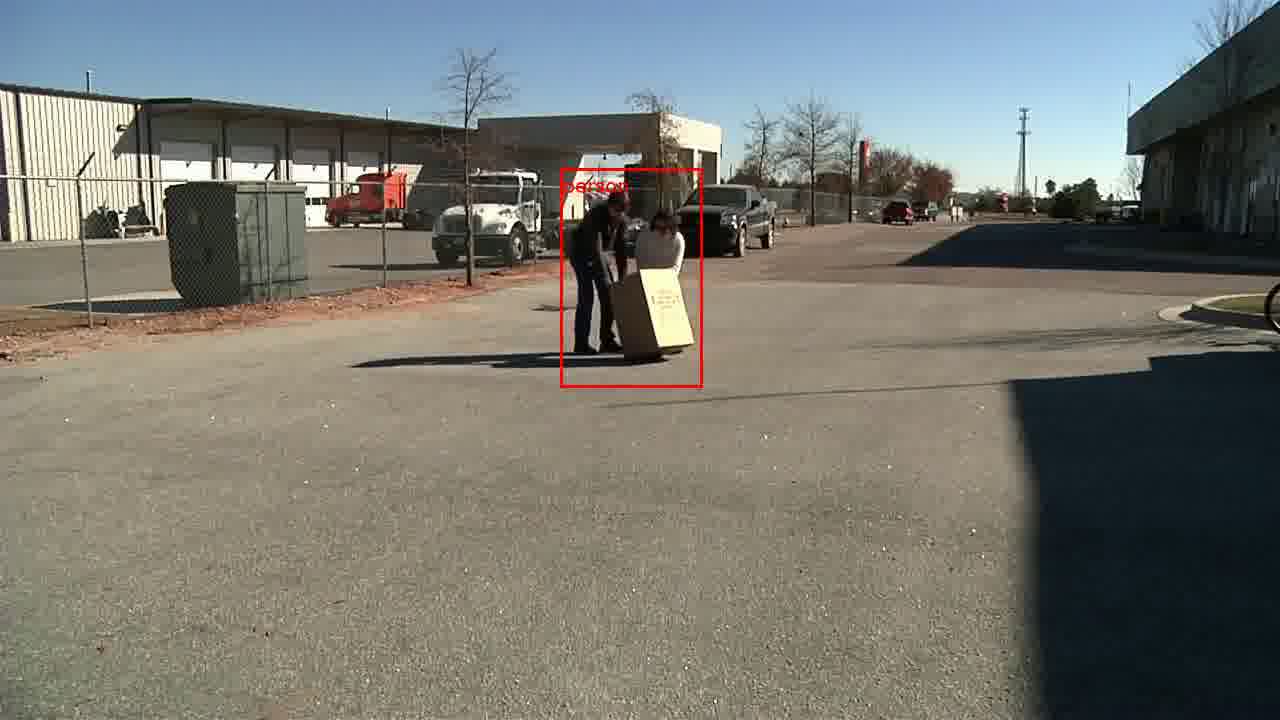}&
      \includegraphics[width=0.24\textwidth]{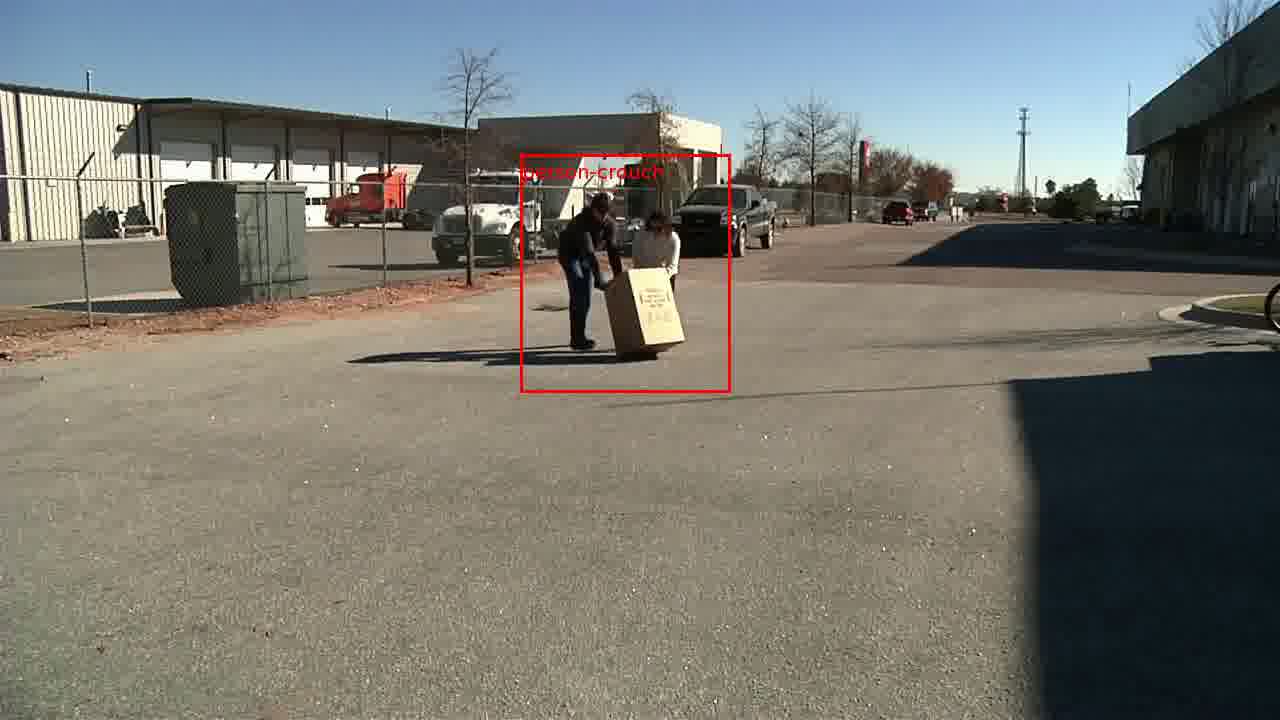}&
      \includegraphics[width=0.24\textwidth]{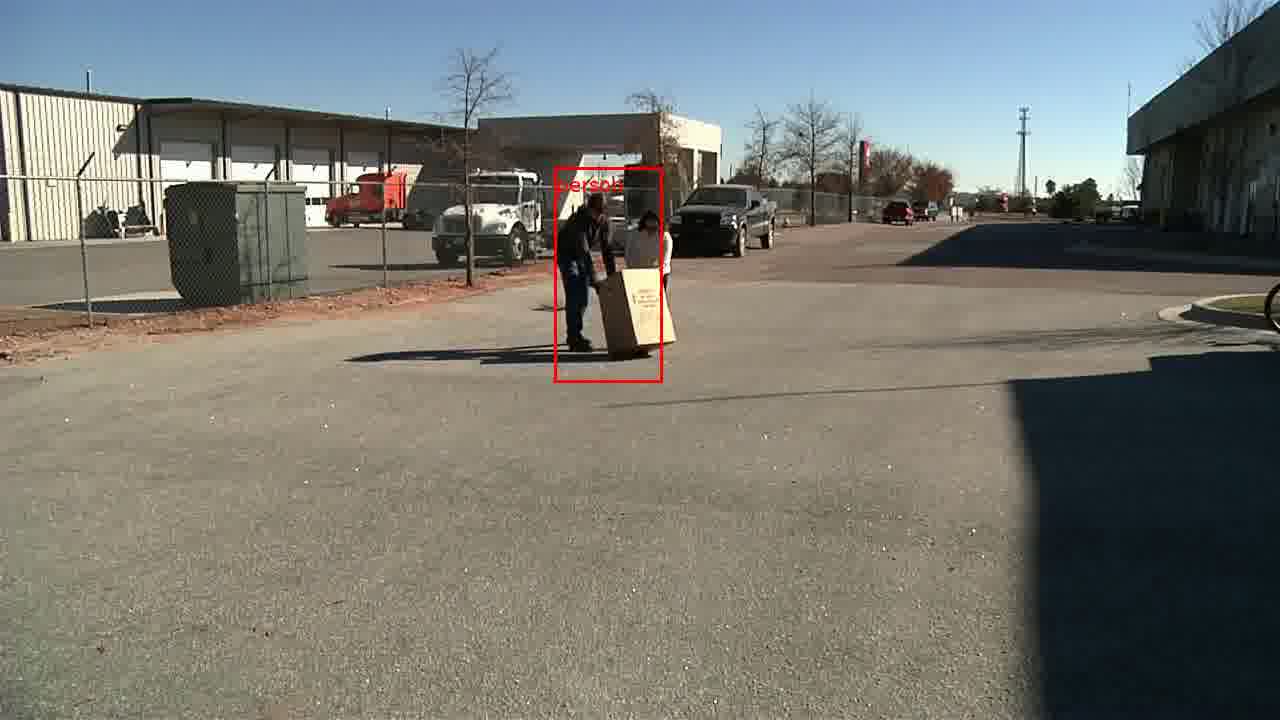}&
      \includegraphics[width=0.24\textwidth]{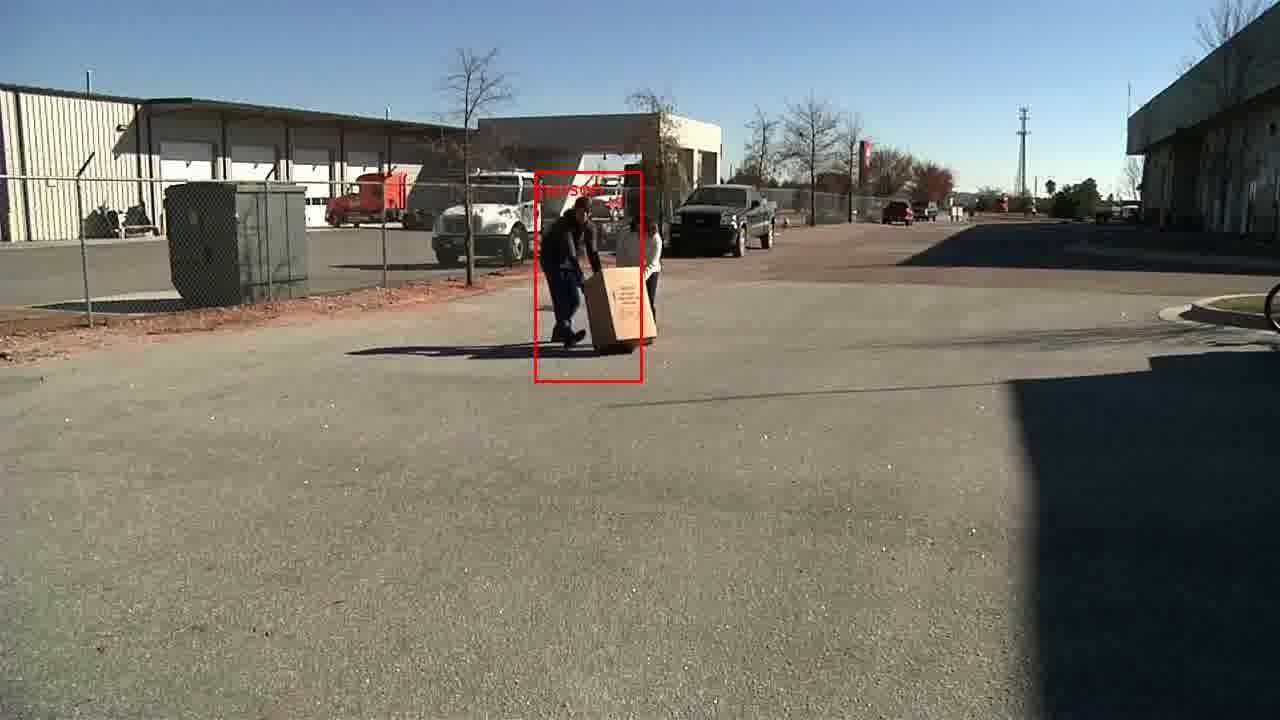}\\
      \multicolumn{4}{c}{The narrow person snatched an object from something.}\\
      \includegraphics[width=0.24\textwidth]{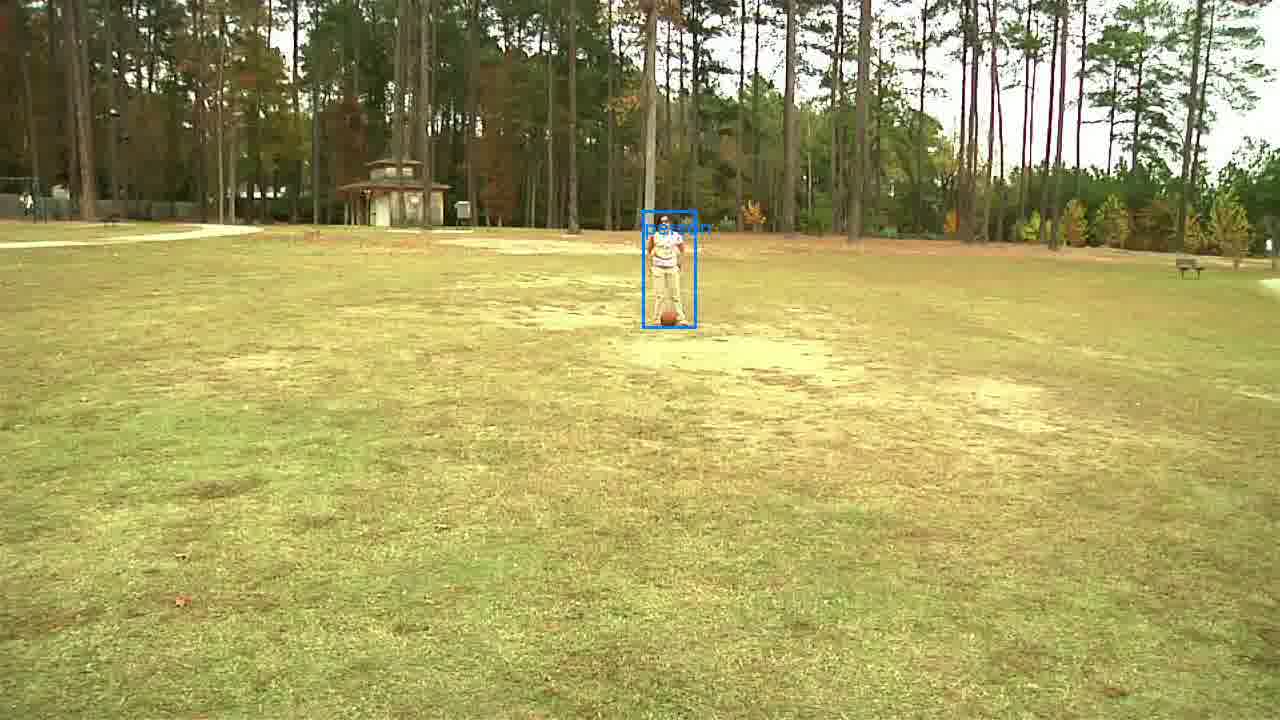}&
      \includegraphics[width=0.24\textwidth]{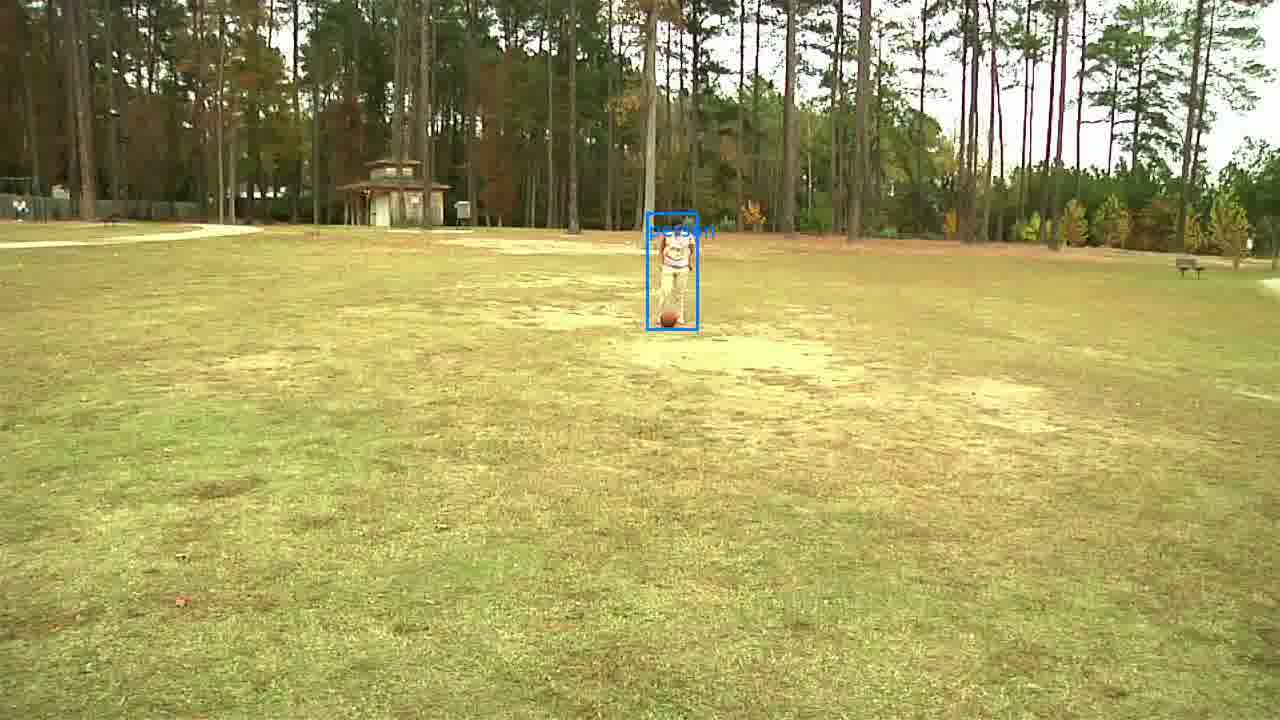}&
      \includegraphics[width=0.24\textwidth]{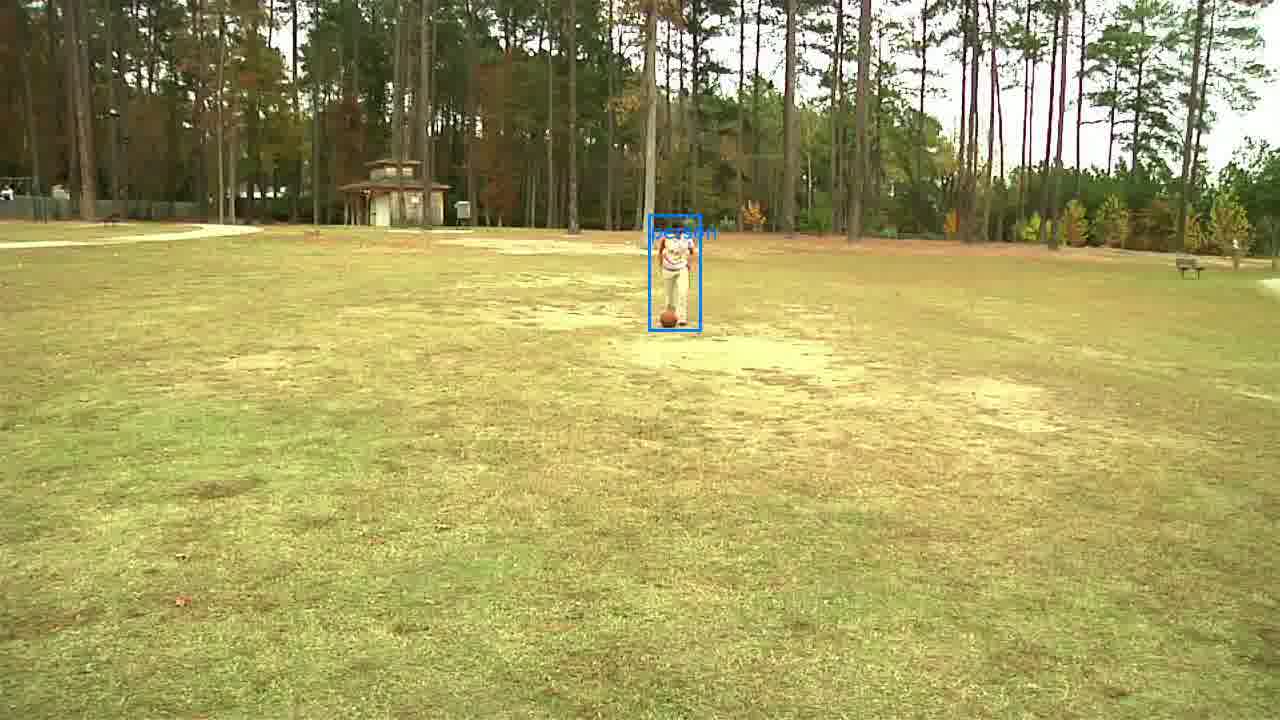}&
      \includegraphics[width=0.24\textwidth]{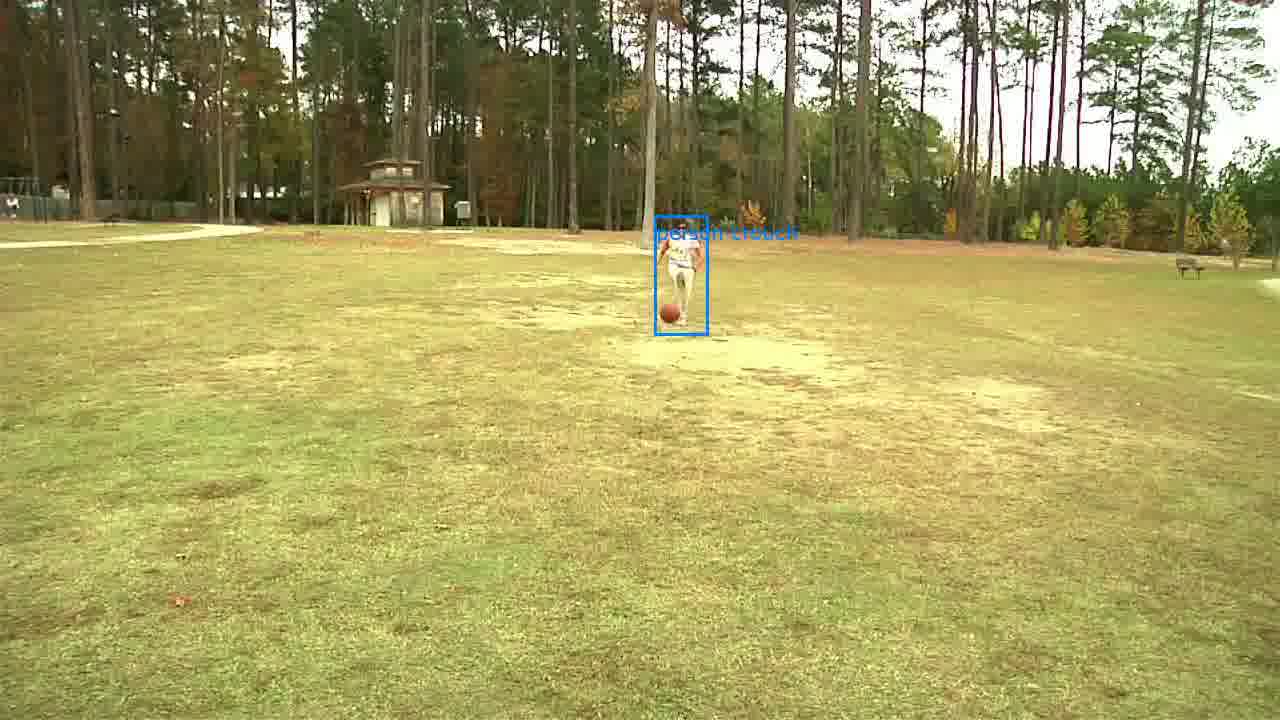}\\
      \includegraphics[width=0.24\textwidth]{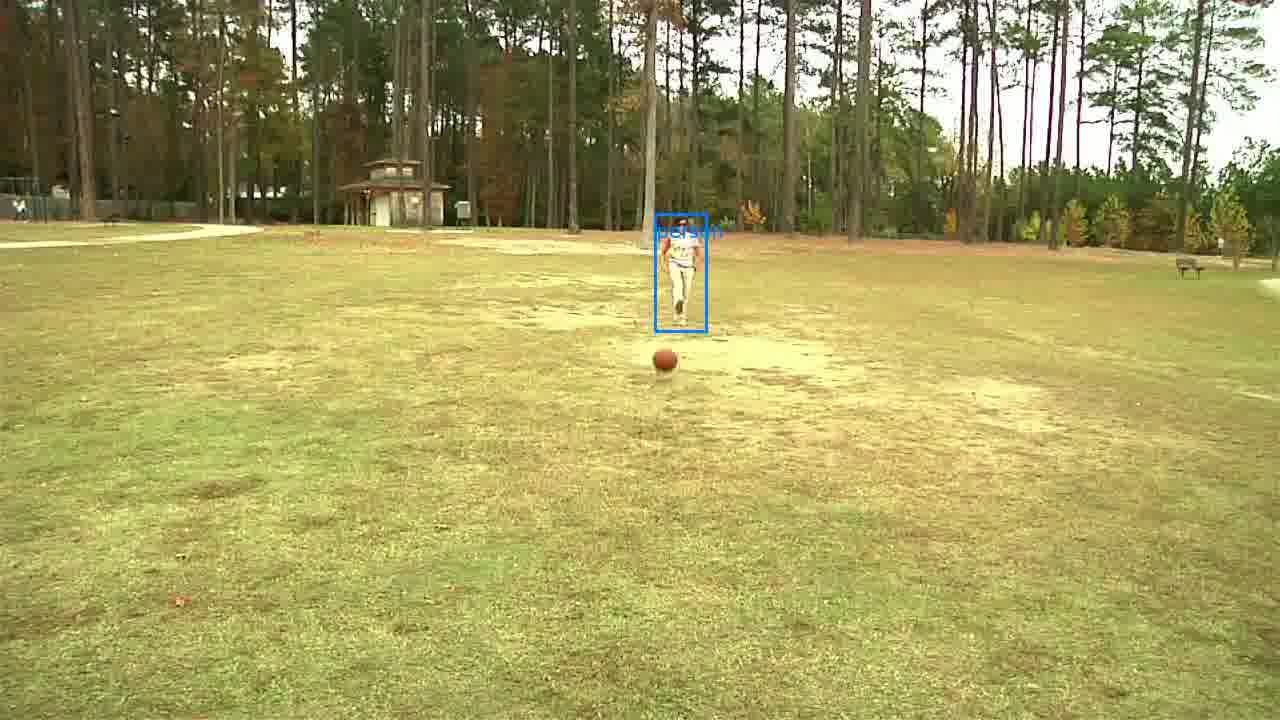}&
      \includegraphics[width=0.24\textwidth]{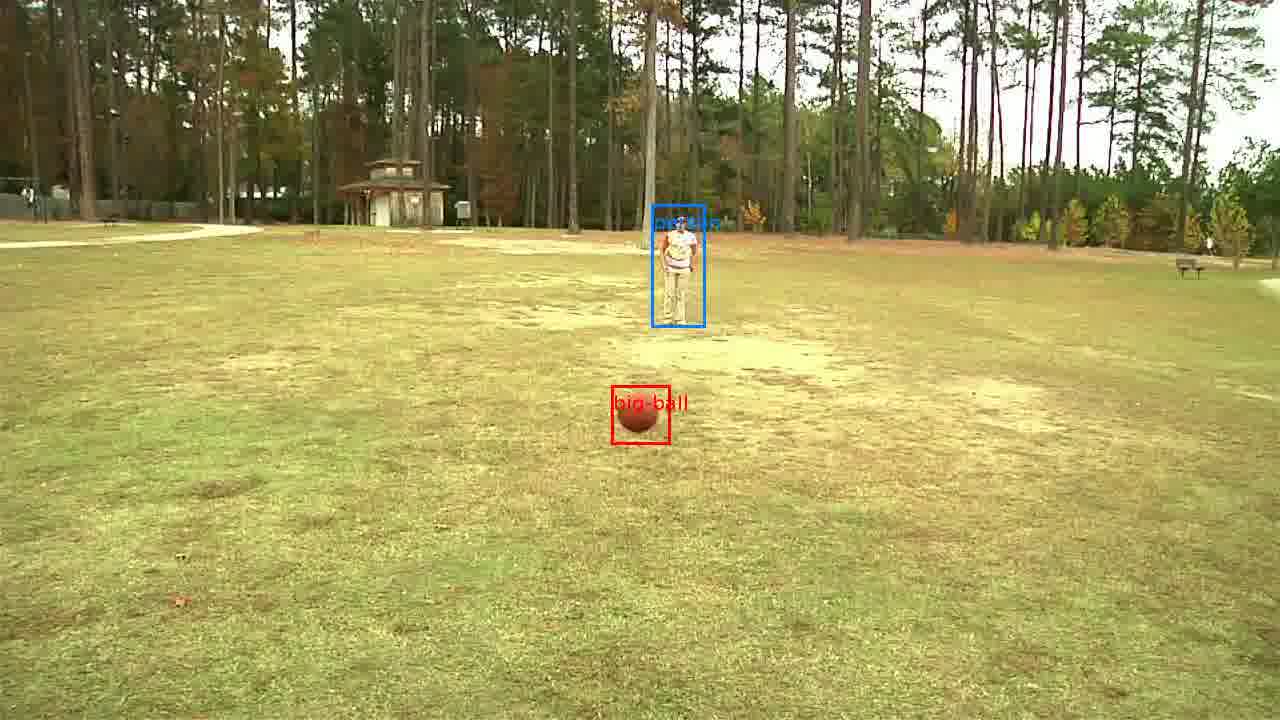}&
      \includegraphics[width=0.24\textwidth]{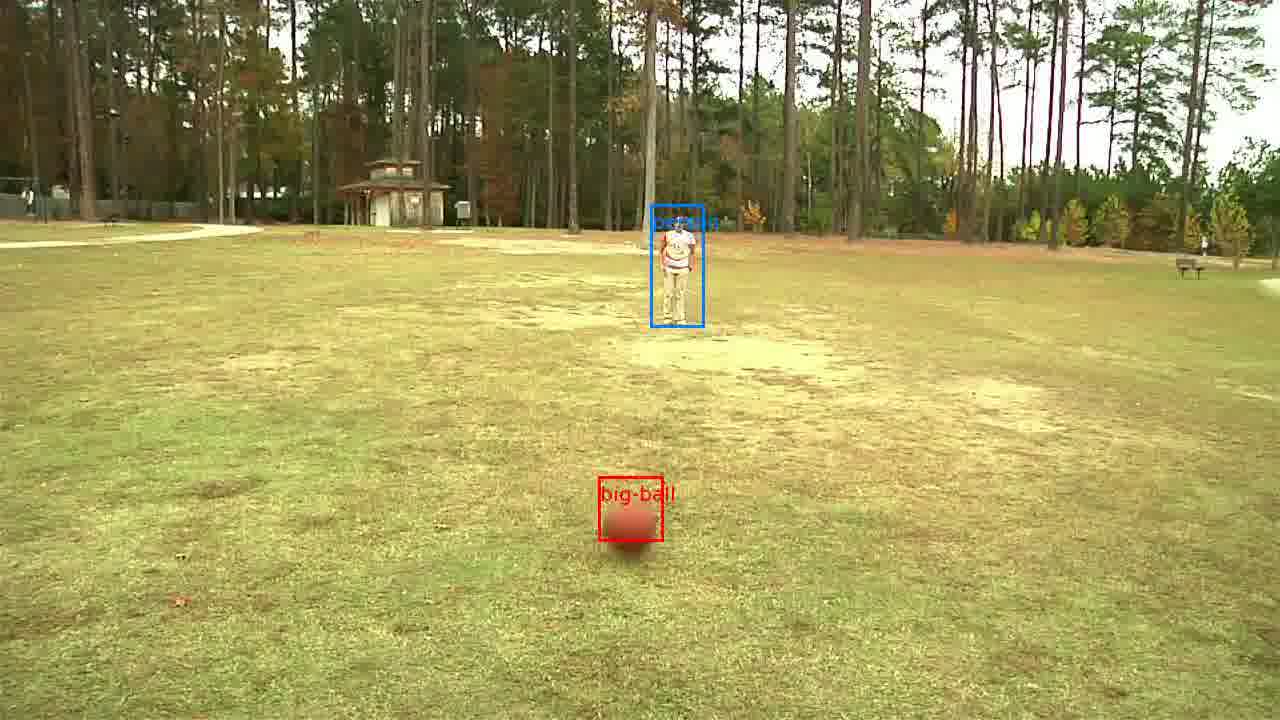}&
      \includegraphics[width=0.24\textwidth]{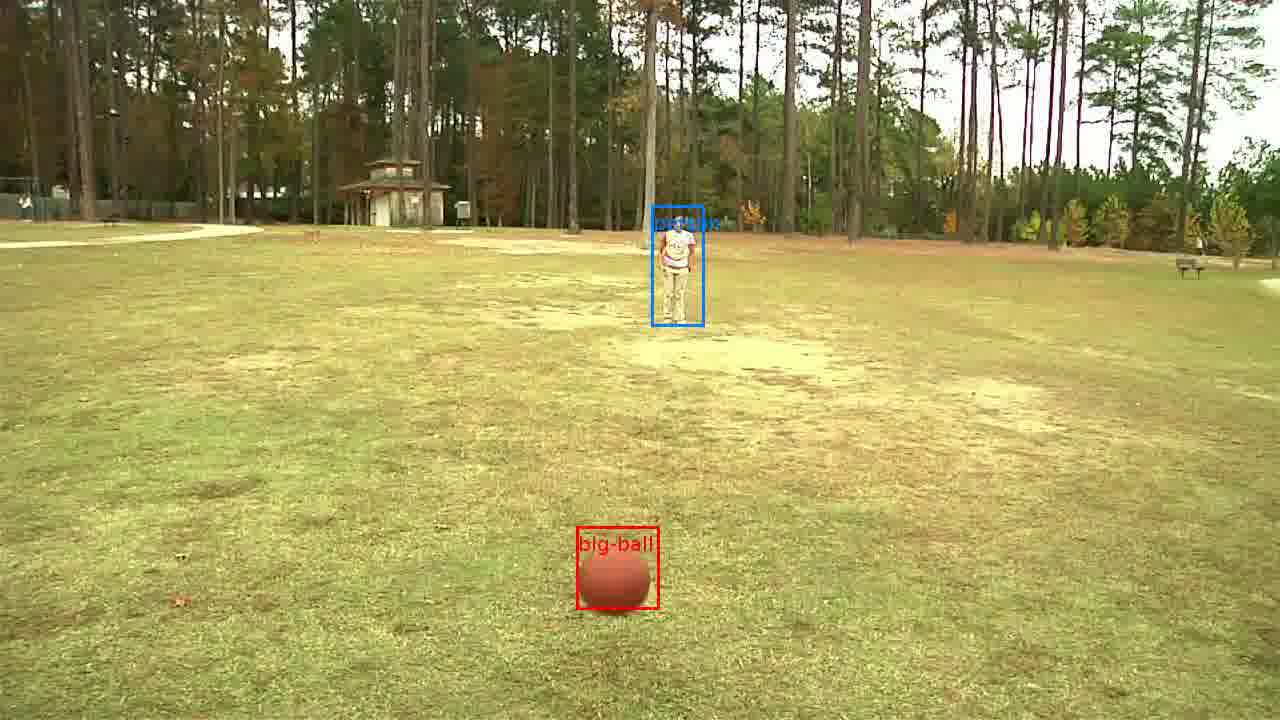}\\
      \multicolumn{4}{c}{The upright person hit the big ball.}
    \end{tabular}
  \end{center}
  \caption{Key frames from four videos in our test set along with the sentence
    generated for the most-likely action class.}
  \label{fig:results}
\end{figure*}

We used the HMMs generated for Experiment~III to compute likelihoods for each
video in our test set paired with each of the 48 action classes.
For each video, we generated sentences corresponding to the three most-likely
action classes.
Fig.~\ref{fig:results} shows key frames from four videos in our test set along
with the sentence generated for the most-likely action class.
Human judges rated each video-sentence pair to assess whether the sentence was
true of the video and whether it described a salient event depicted in that
video.
26.7\% (601/2247) of the video-sentence pairs were deemed to be true and
7.9\% (178/2247) of the video-sentence pairs were deemed to be salient.
When restricting consideration to only the sentence corresponding to the single
most-likely action class for each video, 25.5\% (191/749) of the
video-sentence pairs were deemed to be true and 8.4\% (63/749) of the
video-sentence pairs were deemed to be salient.
Finally, for 49.4\% (370/749) of the videos at least one of the three
generated sentences was deemed true and for 18.4\% (138/749) of the videos
at least one of the three generated sentences was deemed salient.

\section{Conclusion}

Integration of Language and Vision \citep{AAAI2011, HLT-NAACL2003,
  NIPS2011, AAAI1994, McKevitt1995} and
recognition of action in video \citep{Blank2005, Laptev2008, Liu2009,
  Rodriguez2008, Schuldt2004, Siskind1996, Starner98, Wang2009, Xu2002, Xu2005}
have been of considerable interest for a long time.
There has also been work on generating sentential descriptions of static images
\citep{Farhadi2009, Kulkarni2011, Yao2010}.
Yet we are unaware of any prior work that generates as rich sentential
video descriptions as we describe here.
Producing such rich descriptions requires determining event participants, the
mapping of such participants to roles in the event, and their motion and
properties.
This is incompatible with common approaches to event recognition, such as
spatiotemporal bags of words, spatiotemporal volumes, and tracked feature points
that cannot determine such information.
The approach presented here recovers the information needed to generate rich
sentential descriptions by using detection-based tracking and a body-posture
codebook.
We demonstrated the efficacy of this approach on a corpus 749 videos.
\section*{Acknowledgments}
This work was supported, in part, by NSF grant CCF-0438806, by the Naval
Research Laboratory under Contract Number N00173-10-1-G023, by the Army
Research Laboratory accomplished under Cooperative Agreement Number
W911NF-10-2-0060, and by computational resources provided by Information
Technology at Purdue through its Rosen Center for Advanced Computing.
Any views, opinions, findings, conclusions, or recommendations contained or
expressed in this document or material are those of the author(s) and do not
necessarily reflect or represent the views or official policies, either
expressed or implied, of NSF, the Naval Research Laboratory, the Office of
Naval Research, the Army Research Laboratory, or the U.S. Government.
The U.S. Government is authorized to reproduce and distribute reprints for
Government purposes, notwithstanding any copyright notation herein.

\bibliographystyle{plainnat}
\bibliography{arxiv2012b}

\begin{thebibliography}{36}
\providecommand{\natexlab}[1]{#1}
\providecommand{\url}[1]{\texttt{#1}}
\expandafter\ifx\csname urlstyle\endcsname\relax
  \providecommand{\doi}[1]{doi: #1}\else
  \providecommand{\doi}{doi: \begingroup \urlstyle{rm}\Url}\fi

\bibitem[Aloimonos et~al.(2011)Aloimonos, Fadiga, Metta, and Pastra]{AAAI2011}
Y.~Aloimonos, L.~Fadiga, G.~Metta, and K.~Pastra, editors.
\newblock \emph{AAAI Workshop on Language-Action Tools for Cognitive Artificial
  Agents: Integrating Vision, Action and Language}, 2011.

\bibitem[Andriluka et~al.(2008)Andriluka, Roth, and Schiele]{Andriluka2008}
M.~Andriluka, S.~Roth, and B.~Schiele.
\newblock People-tracking-by-detection and people-detection-by-tracking.
\newblock In \emph{CVPR}, pages 1--8, 2008.

\bibitem[Barzialy et~al.(2003)Barzialy, Reiter, and Siskind]{HLT-NAACL2003}
R.~Barzialy, E.~Reiter, and J.M. Siskind, editors.
\newblock \emph{HLT-NAACL Workshop on Learning Word Meaning from Non-Linguistic
  Data}, 2003.

\bibitem[Blank et~al.(2005)Blank, Gorelick, Shechtman, Irani, and
  Basri]{Blank2005}
M.~Blank, L.~Gorelick, E.~Shechtman, M.~Irani, and R.~Basri.
\newblock Actions as space-time shapes.
\newblock In \emph{ICCV}, pages 1395--402, 2005.

\bibitem[Bregler(1997)]{Bregler1997}
Christoph Bregler.
\newblock Learning and recognizing human dynamics in video sequences.
\newblock In \emph{CVPR}, 1997.

\bibitem[Darrell et~al.(2011)Darrell, Mooney, and Saenko]{NIPS2011}
T.~Darrell, R.~Mooney, and K.~Saenko, editors.
\newblock \emph{NIPS Workshop on Integrating Language and Vision}, 2011.

\bibitem[Farhadi et~al.(2009)Farhadi, Endres, Hoiem, and Forsyth]{Farhadi2009}
Ali Farhadi, Ian Endres, Derek Hoiem, and David Forsyth.
\newblock Describing objects by their attributes.
\newblock In \emph{CVPR}, 2009.

\bibitem[Felzenszwalb et~al.(2010{\natexlab{a}})Felzenszwalb, Girshick, and
  McAllester]{Felzenszwalb2010b}
P.~F. Felzenszwalb, R.~B. Girshick, and D.~McAllester.
\newblock Cascade object detection with deformable part models.
\newblock In \emph{CVPR}, 2010{\natexlab{a}}.

\bibitem[Felzenszwalb et~al.(2010{\natexlab{b}})Felzenszwalb, Girshick,
  McAllester, and Ramanan]{Felzenszwalb2010a}
P.~F. Felzenszwalb, R.~B. Girshick, D.~McAllester, and D.~Ramanan.
\newblock Object detection with discriminatively trained part based models.
\newblock \emph{PAMI}, 32\penalty0 (9), 2010{\natexlab{b}}.

\bibitem[Gavrila and Davis(1995)]{GavrilaD95}
D.~M. Gavrila and L.~S. Davis.
\newblock Towards 3-d model-based tracking and recognition of human movement.
\newblock In \emph{International Workshop on Face and Gesture Recognition},
  1995.

\bibitem[Grice(1975)]{Grice1975}
H.~P. Grice.
\newblock Logic and conversation.
\newblock In P.~Cole and J.~L. Morgan, editors, \emph{Syntax and Semantics 3:
  Speech Acts}, pages 41--58. Academic Press, 1975.

\bibitem[Ikizler-Cinibis and Sclaroff(2010)]{Ikizler2010}
Nazli Ikizler-Cinibis and Stan Sclaroff.
\newblock Object, scene and actions: Combining multiple features for human
  action recognition.
\newblock In \emph{ECCV}, pages 494--507, 2010.

\bibitem[Jackendoff(1983)]{Jackendoff83}
Ray Jackendoff.
\newblock \emph{Semantics and Cognition}.
\newblock MIT Press, Cambridge, {MA}, 1983.

\bibitem[Kulkarni et~al.(2011)Kulkarni, Premraj, Dhar, Li, Choi, Berg, and
  Berg]{Kulkarni2011}
Girish Kulkarni, Visruth Premraj, Sagnik Dhar, Siming Li, Yejin Choi,
  Alexander~C. Berg, and Tamara~L. Berg.
\newblock Baby talk: Understanding and generating simple image descriptions.
\newblock In \emph{CVPR}, pages 1601--8, 2011.

\bibitem[Laptev et~al.(2007)Laptev, Caputo, Schuldt, and Lindeberg]{Laptev2007}
I.~Laptev, B.~Caputo, C.~Schuldt, and T.~Lindeberg.
\newblock Local velocity-adapted motion events for spatio-temporal recognition.
\newblock \emph{CVIU}, 108\penalty0 (3):\penalty0 207--29, 2007.

\bibitem[Laptev et~al.(2008)Laptev, Marszalek, Schmid, and
  Rozenfeld]{Laptev2008}
I.~Laptev, M.~Marszalek, C.~Schmid, and B.~Rozenfeld.
\newblock Learning realistic human actions from movies.
\newblock In \emph{CVPR}, 2008.

\bibitem[Liu et~al.(2009)Liu, Luo, and Shah]{Liu2009}
J.~Liu, J.~Luo, and M.~Shah.
\newblock Recognizing realistic actions from videos ``in the wild''.
\newblock In \emph{CVPR}, pages 1996--2003, 2009.

\bibitem[McKevitt(1994)]{AAAI1994}
P.~McKevitt, editor.
\newblock \emph{AAAI Workshop on Integration of Natural Language and Vision
  Processing}, 1994.

\bibitem[McKevitt(1995--1996)]{McKevitt1995}
P.~McKevitt, editor.
\newblock \emph{Integration of Natural Language and Vision Processing}, volume
  I--IV.
\newblock Kluwer, Dordrecht, 1995--1996.

\bibitem[Niebles et~al.(2008)Niebles, Wang, and Fei-Fei]{Niebles2008}
J.~C. Niebles, H.~Wang, and L.~Fei-Fei.
\newblock Unsupervised learning of human action categories using
  spatial-temporal words.
\newblock \emph{IJCV}, 79\penalty0 (3):\penalty0 299--318, 2008.

\bibitem[Otsu(1979)]{Otsu1979}
N.~Otsu.
\newblock A threshold selection method from gray-level histograms.
\newblock \emph{{IEEE} Trans. on Systems, Man and Cybernetics}, 9\penalty0
  (1):\penalty0 62--6, 1979.
\newblock ISSN 0018-9472.

\bibitem[Pinker(1989)]{Pinker89}
Steven Pinker.
\newblock \emph{Learnability and Cognition}.
\newblock MIT Press, Cambridge, {MA}, 1989.

\bibitem[Rodriguez et~al.(2008)Rodriguez, Ahmed, and Shah]{Rodriguez2008}
M.~D. Rodriguez, J.~Ahmed, and M.~Shah.
\newblock Action {MACH}: A spatio-temporal maximum average correlation height
  filter for action recognition.
\newblock In \emph{CVPR}, 2008.

\bibitem[Schuldt et~al.(2004)Schuldt, Laptev, and Caputo]{Schuldt2004}
C.~Schuldt, I.~Laptev, and B.~Caputo.
\newblock Recognizing human actions: A local {SVM} approach.
\newblock In \emph{ICPR}, pages 32--6, 2004.
\newblock ISBN 0-7695-2128-2.

\bibitem[Scovanner et~al.(2007)Scovanner, Ali, and Shah]{Scovanner2007}
P.~Scovanner, S.~Ali, and M.~Shah.
\newblock A 3-dimensional {SIFT} descriptor and its application to action
  recognition.
\newblock In \emph{International Conference on Multimedia}, pages 357--60,
  2007.

\bibitem[Shi and Tomasi(1994)]{shi1994}
J.~Shi and C.~Tomasi.
\newblock Good features to track.
\newblock In \emph{CVPR}, pages 593--600, 1994.

\bibitem[Sigal et~al.(2010)Sigal, Balan, and Black]{Sigal2010}
L.~Sigal, A.~Balan, and M.~J. Black.
\newblock {HumanEva}: Synchronized video and motion capture dataset and
  baseline algorithm for evaluation of articulated human motion.
\newblock \emph{IJCV}, 87\penalty0 (1-2):\penalty0 4--27, 2010.

\bibitem[Siskind and Morris(1996)]{Siskind1996}
J.~M. Siskind and Q.~Morris.
\newblock A maximum-likelihood approach to visual event classification.
\newblock In \emph{ECCV}, pages 347--60, 1996.

\bibitem[Starner et~al.(1998)Starner, Weaver, and Pentland]{Starner98}
Thad Starner, Joshua Weaver, and Alex Pentland.
\newblock Real-time {A}merican sign language recognition using desk and
  wearable computer based video.
\newblock \emph{PAMI}, 20\penalty0 (12):\penalty0 1371--5, 1998.

\bibitem[Tomasi and Kanade(1991)]{tomasi1991}
C.~Tomasi and T.~Kanade.
\newblock Detection and tracking of point features.
\newblock Technical Report CMU-CS-91-132, Carnegie Mellon University, 1991.

\bibitem[Viterbi(1971)]{Viterbi1971}
A.~J. Viterbi.
\newblock Convolutional codes and their performance in communication systems.
\newblock \emph{{IEEE} Trans. on Communication}, 19:\penalty0 751--72, 1971.

\bibitem[Wang and Mori(2009)]{Wang2009}
Y.~Wang and G.~Mori.
\newblock Human action recognition by semilatent topic models.
\newblock \emph{PAMI}, 31\penalty0 (10):\penalty0 1762--74, 2009.
\newblock ISSN 0162-8828.

\bibitem[Xu et~al.(2002)Xu, Ma, Zhang, and Yang]{Xu2002}
Gu~Xu, Yu-Fei Ma, HongJiang Zhang, and Shiqiang Yang.
\newblock Motion based event recognition using {HMM}.
\newblock In \emph{ICPR}, volume~2, 2002.

\bibitem[Xu et~al.(2005)Xu, Ma, Zhang, and Yang]{Xu2005}
Gu~Xu, Yu-Fei Ma, HongJiang Zhang, and Shi-Qiang Yang.
\newblock An {HMM}-based framework for video semantic analysis.
\newblock \emph{IEEE Trans. Circuits Syst. Video Techn.}, 15\penalty0
  (11):\penalty0 1422--33, 2005.

\bibitem[Yang and Ramanan(2011)]{Yang2011}
Y.~Yang and D.~Ramanan.
\newblock Articulated pose estimation using flexible mixtures of parts.
\newblock In \emph{CVPR}, 2011.

\bibitem[Yao et~al.(2010)Yao, Yang, Lin, Lee, and Zhu]{Yao2010}
B.~Z. Yao, Xiong Yang, Liang Lin, Mun~Wai Lee, and Song-Chun Zhu.
\newblock I2t: Image parsing to text description.
\newblock \emph{Proceedings of the IEEE}, 98\penalty0 (8):\penalty0 1485--1508,
  2010.

\end{thebibliography}

\end{document}